%% file: CrossPoints.tex
\documentclass[lettersize,journal]{IEEEtran}
\usepackage{amsmath,amsfonts}
\usepackage{xcolor}
\input{def}

\usepackage{array}
\usepackage[caption=false,font=normalsize,labelfont=sf,textfont=sf]{subfig}
\usepackage{textcomp}
\usepackage{stfloats}
\usepackage{url}
\usepackage{verbatim}
\usepackage{graphicx}
\usepackage{cite}
\usepackage{ifpdf}
\usepackage[ruled,vlined]{algorithm2e}
\usepackage{amssymb}
\usepackage{booktabs} 
\usepackage{bbding}
\usepackage{pifont}
\begin{document}

\title{Cross-view Domain Generalization via Geometric Consistency for LiDAR Semantic Segmentation}

\author{Jindong~Zhao, Yuan~Gao, Yang~Xia, Sheng Nie, Jun Yue, Weiwei Sun, Shaobo~Xia

\thanks{This work was supported in part by the National Natural Science Foundation of China under Grant 62575039, Grant 62535017, Grant 62575311, and in part by the Hunan Provincial Natural Science Foundation of China under Grant 2025JJ40059. (Jindong Zhao and Yuan Gao contributed equally to this work) (Corresponding author: Shaobo Xia)}
\thanks{Jindong Zhao, Yang Xia and Shaobo Xia are with the School of Aeronautic Engineering, Changsha University of Science and Technology, Hunan, China, 410004. (E-mail: zkintom@gmail.com, xiayang0302@gmail.com, shaoboxia2020@gmail.com)}
\thanks{Yuan Gao and Sheng Nie are with the Aerospace Information Research Institute, Chinese Academy of Sciences, Beijing, China, 100094. (e-mail: gaoyuan222@mails.ucas.ac.cn, niesheng@aircas.ac.cn).}
\thanks{Jun Yue is with the School of Automation, Central South University, Changsha 410083, China (e-mail: junyue@csu.edu.cn).}
\thanks{Weiwei Sun is with the University of British Columbia. (weiwei.sun3@gmail.com)}
}



\maketitle

\input{sec/0_abstract}


\input{sec/1_intro}

\input{sec/2_related}

\input{sec/3_method}

\input{sec/4_experiments}

\input{sec/5_conclusions.tex}



\ifCLASSOPTIONcaptionsoff
  \newpage
\fi

\bibliographystyle{IEEEtran}
\bibliography{IEEEabrv,cross.bib}

\end{document}

%% file: def.tex
\usepackage{pgfplots}
\usepackage{multirow}  
\usepackage{arydshln} 
\usepackage{makecell}   
\usepackage{booktabs}   
\usepackage{tikz}
\usepackage{mathtools}
\usetikzlibrary{matrix, positioning,arrows,arrows.meta,calc,decorations.pathreplacing,decorations.text,spy}
\pgfplotsset{compat=1.18}

\tikzset{
  img/.style={
    inner sep=0pt,     
    outer sep=0pt,     
    rectangle,
    align=center} 
}

\usepackage[colorlinks,linkcolor=blue,citecolor=red]{hyperref} 

\usepackage{soul}

\usepackage{dsfont}

\usepackage{blindtext}

\usepackage{enumitem}
\setlist[itemize]{noitemsep,leftmargin=*,topsep=.05in}
\setlist[enumerate]{noitemsep,leftmargin=*,topsep=.05in}

\def \customparskip {0.5em}
\renewcommand{\paragraph}[1]{\vspace{\customparskip}\noindent\textbf{#1}.}



%% file: sec/0_abstract.tex
\begin{abstract}
Domain-generalized LiDAR semantic segmentation (LSS) seeks to train models on source-domain point clouds that generalize reliably to multiple unseen target domains, which is essential for real-world LiDAR applications. However, existing approaches assume similar acquisition views (e.g., vehicle-mounted) and struggle in cross-view scenarios, where observations differ substantially due to viewpoint-dependent structural incompleteness and non-uniform point density. Accordingly, we formulate cross-view domain generalization for LiDAR semantic segmentation and propose a novel framework, termed CVGC (Cross-View Geometric Consistency). Specifically, we introduce a cross-view geometric augmentation module that models viewpoint-induced variations in visibility and sampling density, generating multiple cross-view observations of the same scene. Subsequently, a geometric consistency module enforces consistent semantic and occupancy predictions across geometrically augmented point clouds of the same scene. Extensive experiments on six public LiDAR datasets establish the first systematic evaluation of cross-view domain generalization for LiDAR semantic segmentation, demonstrating that CVGC consistently outperforms state-of-the-art methods when generalizing from a single source domain to multiple target domains with heterogeneous acquisition viewpoints. The source code will be publicly available at~\url{https://github.com/KintomZi/CVGC-DG}

\end{abstract}

%% file: sec/1_intro.tex
\section{Introduction}
\IEEEPARstart{S}{emantic} segmentation of LiDAR point clouds is a fundamental task for 3D scene understanding and has a wide range of applications, including autonomous driving, robotics, and earth observation~\cite{geiger2012we,lafarge2012creating,pan2024pin, gao2025lidar}. Deep neural networks ~\cite{choy20194d,guo2020deep,xu2025frnet} have achieved remarkable success on this task when training and test data follow similar distributions. However, real-world scenarios frequently violate this assumption. When deployed across diverse environments, LiDAR semantic segmentation (LSS) models inevitably encounter substantial domain shifts caused by variations in scene layouts (e.g., urban vs. rural environments) and acquisition protocols (e.g., sensor specifications, scanning patterns and platform configuration). Such domain shifts severely degrade the performance compared to in-domain performance. A common strategy is to collect target-domain data for domain adaptation (DA), but in practice acquiring and annotating such data is often prohibitively expensive or infeasible. This practical limitation naturally motivates the domain generalization (DG) paradigm~\cite{zhou2022domain,wang2022generalizing}, where models are trained solely on source domains and are expected to generalize directly to unseen target domains.

\begin{figure}[t]
    \centering
    \includegraphics[width=\columnwidth]{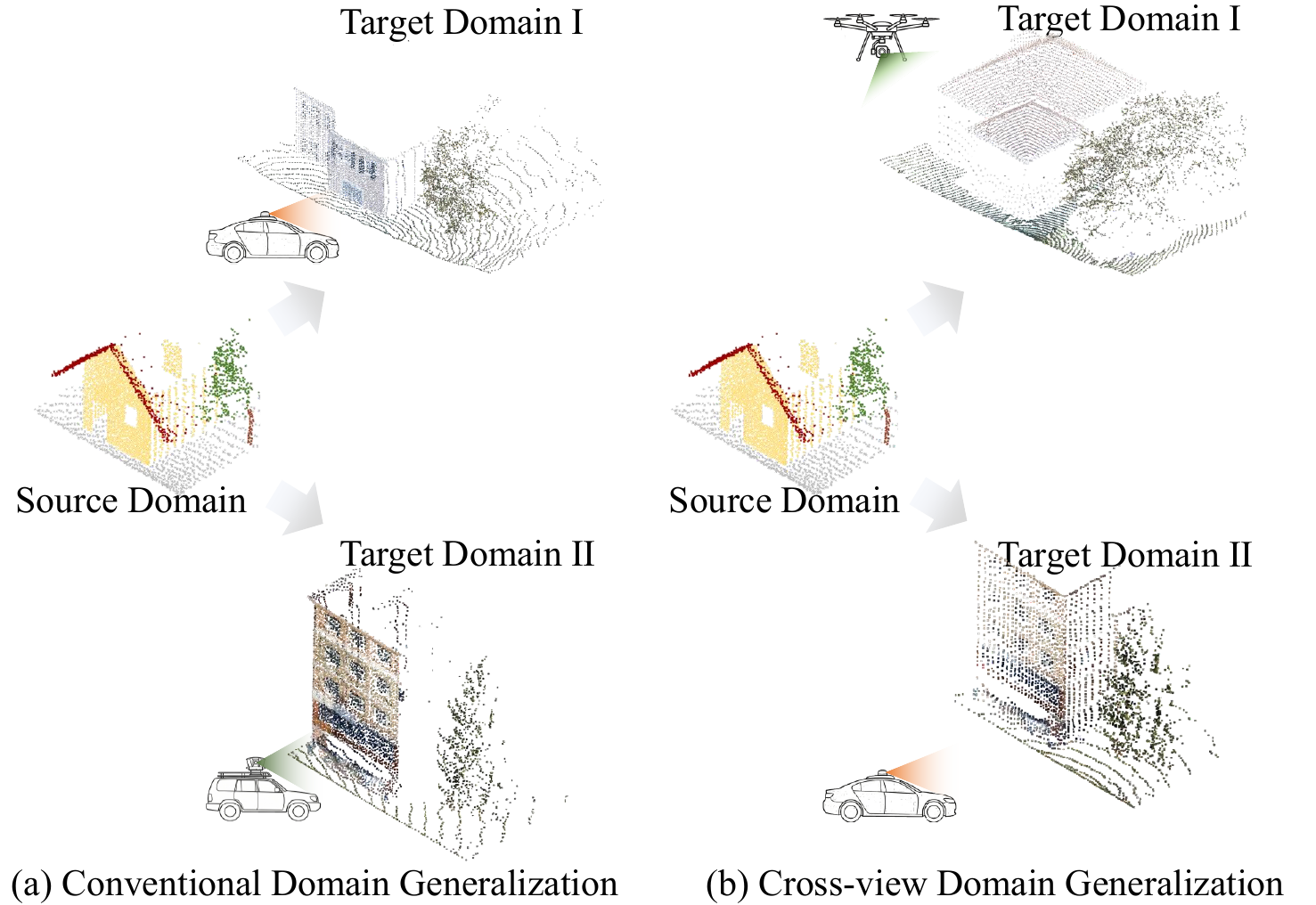}
    \caption{\textbf{Challenges in cross-view domain generalization.} (a) Conventional DG assumes similar sensing viewpoints across domains, whereas (b) Cross-view DG involves drastically different acquisition geometries, leading to severe density and visibility shifts. We address this challenge by generating view-dependent geometric variants and enforcing consistency across them.
    }
    \label{fig:teaser}
\end{figure}

Conventional DG for LSS~\cite{zhou2022domain,wang2022generalizing,kim2023single,zhao2024unimix} has achieved notable progress under the assumption of comparable sensing viewpoints, enabling models to exploit shared structures such as building facades and road surfaces, as illustrated in Fig.~\ref{fig:teaser}(a). In practice, LiDAR data are acquired from heterogeneous platforms with fundamentally different sensing viewpoints and scanning mechanisms, leading to drastic density and visibility variations, as shown in Fig.~\ref{fig:teaser}(b). With the rapid adoption of flexible-viewpoint platforms such as UAVs, LiDAR data collection has become increasingly diverse, and real-world applications often require joint processing of data from multiple viewpoints, driving a growing demand for multi-view collaboration and platform-agnostic deployment~\cite{shi2025l2rsi,10657359,10373898,Liang_2025_ICCV}. These trends call for cross-view DG for LSS, enabling models trained on one platform to generalize to unseen domains captured from heterogeneous sensing viewpoints~\cite{ren2022adela,truong2024eagle,Liang_2025_ICCV}.

Cross-view DG has been widely explored in image segmentation~\cite{coors2019nova,ren2022adela,klinghoffer2023towards,truong2024eagle}, but remains largely unexplored in LSS. Existing DG methods for LSS are predominantly developed for vehicle-mounted LiDAR, in which similar viewpoints and scanning configurations result in relatively consistent density and occlusion shifts~\cite{kim2023single,zhao2024unimix}. In contrast, cross-view LiDAR involves fundamentally different sensing geometries across airborne, UAV-based, and vehicle-mounted platforms, resulting in drastic discrepancies in point density distribution and structure visibility. For example, vehicle-mounted LiDAR primarily captures vertical structures such as building facades, while airborne LiDAR mainly observes horizontal structures such as rooftops, as shown in Fig.~\ref{fig:teaser} (b). Consequently, existing LiDAR DG methods assume that the geometric structures in source and target domains are inherently comparable, and therefore do not explicitly address the drastic structural discrepancies introduced by cross-view acquisition. To overcome this limitation, we propose a self-supervised geometric consistency framework that does not assume inherent geometric alignability between source and target domains. Instead, it constructs multiple view-dependent variants of the same source scene and explicitly aligns them by enforcing semantic and structural consistency among these variants. In summary, our contributions are threefold:

\begin{enumerate}
  \item We formulate, for the first time, the problem of cross-view domain generalization for LiDAR point cloud semantic segmentation, and propose a novel framework, termed CVGC, to address the fundamental challenges induced by heterogeneous LiDAR acquisition viewpoints.

  \item We introduce a geometric-consistency-driven learning framework that generates multiple cross-view variants of the same scene via geometric augmentation and enforces semantic and occupancy consistency across these observations, enabling robust generalization under structural incompleteness and density variation.

  \item We establish the first comprehensive benchmark for cross-view domain generalization in LiDAR semantic segmentation, spanning six public datasets across airborne, UAV-based, and vehicle-mounted platforms, which reveals the critical impact of acquisition viewpoints and enables state-of-the-art performance with CVGC.
\end{enumerate}

The remainder of this paper is organized as follows. Section II reviews related work. Section III describes the proposed method. Section IV presents experimental evaluations. Section V concludes the paper.

%% file: sec/2_related.tex
\section{Related works}
\subsection{Domain Adaptation for LSS}
LiDAR semantic segmentation aims to assign semantic labels to each 3D point, enabling fine-grained scene understanding for applications such as autonomous driving and remote sensing~\cite{varney2020dales, guo2020deep}. 
However, the data distribution of LiDAR point clouds is strongly shaped by acquisition geometry (viewpoint/altitude), sensor configuration, and scene layout, causing pronounced domain shifts. Domain adaptation (DA) studies this problem by assuming that labeled data from the target domain are accessible during training~\cite{rist2019cross,xiao2022transfer}. Since dense point-wise annotations are expensive, recent works increasingly consider unsupervised domain adaptation (UDA)~\cite{cosmix2023,shaban2023lidar,PrototypeUDA2023,chen2024bridging,luo2025cross}, where only labeled source data and unlabeled target data are available.

Cross-domain mixing is a widely used technique in LiDAR UDA~\cite{xiao2022polarmix,cosmix2023,xiao2023domain}, which explicitly relies on access to target-domain data to construct mixed domains and enhance target-domain performance through self-training.
For example, CoSMix~\cite{saltori2022cosmix} bidirectionally mixes source point clouds with pseudo-labeled target data within a teacher–student framework, enabling effective synthetic-to-real adaptation. PolarMix~\cite{xiao2022polarmix} leverages LiDAR-specific geometry by conducting cut-and-mix operations in polar coordinates. LaserMix\cite{kong2023lasermix} further improves realism by incorporating beam-level priors, including the LiDAR scanning pattern and height distribution. UniMix~\cite{zhao2024unimix} constructs a bridge domain via physical simulation on clear-weather data to connect the source and target domains. It enriches training samples using a mixing strategy that encompasses spatial, intensity, and semantic dimensions, while leveraging a teacher-student framework to enhance robustness under adverse conditions.

Target-domain simulation seeks to approximate target-domain distributions, either in data space or feature space, to improve model generalization to target domains. Yuan et al.~\cite{Yuan_2024_CVPR} introduce a Density-Guided Translator that reduces cross-domain discrepancies by translating LiDAR scans across domains based on density statistics. By producing target-like source scans that emulate real sensor sampling patterns, their method yields more realistic training data and significantly improves synthetic-to-real transfer. Xiao et al.~\cite{xiao2024domain} further propose a confidence-modulated framework that enhances pseudo-label reliability by exploiting point-density cues. Their model enforces prediction consistency between raw and sparsified versions of the same point cloud through density-invariant representations, alleviating performance degradation caused by the strong correlation between point density and domain shift. Luo et al.~\cite{luo2020unsupervised} address cross-scene geometric discrepancies by explicitly aligning height distributions between the source and target domains, thereby reducing structural mismatch in complex urban environments and achieve more reliable feature-level adaptation.

In addition to density-based approaches, geometry-oriented target-domain simulation has also been explored. These methods leverage spatial continuity to generate completed or continuous scene representations for domain alignment~\cite{yi2021complete,boulch2023also,SALUDA2024}. For example, Yi et al.~\cite{yi2021complete} reconstruct dense surfaces from sparse LiDAR scans to enable alignment on more complete geometry, while LiT~\cite{lao2024lit} translates source data into the target format using neural implicit representations and target-domain ray statistics. Rather than explicit reconstruction, SALUDA~\cite{SALUDA2024} enforces an implicit surface constraint via auxiliary occupancy estimation, and Luo et al.~\cite{luo2025cross} project different sensors into a unified canonical space using implicit representations.

In summary, prior work on LSS has largely focused on DA and UDA settings, addressing transfer between datasets with similar sensing viewpoints~\cite{li2026its,WANG2025422,whu3d}. Cross-view scenarios, however, involve fundamentally different platforms and geometries, leading to drastic discrepancies in visibility and spatial sampling. These factors create a far more challenging generalization problem, and existing methods typically suffer significant performance degradation when applied to cross-view settings.

\subsection{Domain Generalization for LSS}
DA and UDA require target-domain data during training, which is often infeasible in practical deployments. DG, by contrast, learns solely from labeled source domains and aims to generalize to unseen domains, offering a more practical and scalable solution for LSS. Existing DG methods are commonly grouped into three categories \cite{zhou2022domain,wang2022generalizing}: data manipulation, which augments the training distribution (e.g., Mixup \cite{zhang2018mixup}); representation learning, which seeks domain-invariant features via discrepancy minimization (e.g., adversarial alignment \cite{li2018domain}); and learning strategies, which improve generalization by altering the training paradigm, such as meta-learning \cite{li2018learning}. Although domain generalization has been extensively studied in 2D images, the irregular, sparse, and sensor-dependent nature of 3D point clouds introduces fundamentally different geometric domain shifts, making image-based DG methods difficult to apply directly to LSS.

Research on DG for LSS is currently in its early stages. For data manipulation methods, Espadinha et al. \cite{espadinha2021lidar} enhance simulated data by modeling and injecting noise and point dropout based on real data, thereby making the simulated data closer to the real-world distribution. Xiao et al. \cite{xiao20233d} design an augmentation pipeline tailored to adverse weather, which synthesizes LiDAR data under diverse meteorological conditions. UniMix \cite{zhao2024unimix} further introduces a physically interpretable simulation process to construct an intermediate ``bridge” domain that reduces the domain gap caused by weather changes.  Regarding representation learning, Sanchez et al. \cite{Sanchez_2023_ICCV} leverage geometric accumulation across multiple point cloud frames and temporal label propagation to construct more stable and density-invariant representations. Kim et al. \cite{kim2023single} simulate sparsity variation via random beam sub-sampling and introduce sparse-invariant feature and semantic correlation consistency constraints to preserve both high-dimensional features and semantic structure across domains.  Follow-up work \cite{kim2024rethinking} identifies point density as a key factor underlying cross-sensor discrepancies and introduces a density discriminative feature embedding module to explicitly model density changes arising from different beam configurations. Additionally, Kim et al. \cite{kim2024density} interpret convolutional kernel occupancy patterns as geometric descriptors and incorporate a voxel-level self-supervised density prediction task to strengthen the model’s awareness of density variations. Sanchez et al. \cite{COLA_2025_Sanchez} achieve semantic-level cross-domain alignment by remapping fine-grained labels from multiple datasets to a common coarse label set based on semantic similarity.

A single augmentation scheme or isolated consistency constraint is often inadequate for addressing the complex domain shifts in real-world applications. Consequently, hybrid strategies that combine data processing with representation learning have emerged as a more effective solution. He et al. \cite{he2024domain} expand the feature distribution based on mean–variance statistics to capture target-domain diversity, and propose momentum introspective learning, which introspects and refines prediction discrepancies to generalize to unseen domains. Their subsequent work \cite{he2025domain} propose geometric consistency learning by simulating varying acquisition conditions and enforcing the stability of class-level geometric embeddings. In addition, Li et al. \cite{li2023bev} exploit cross-modal BEV representations to extract more domain-invariant features. Hegde et al. \cite{hegde2025multimodal} combine supervised contrastive learning to achieve alignment within a multimodal feature space that fuses image and point cloud data. Saltori et al. \cite{saltori2023walking} employ 2D BEV semantic prediction as an auxiliary task to guide the network in learning semantic features that are invariant to viewpoint and sensor modality.

In summary, existing DG methods for LiDAR semantic segmentation are vehicle-centric and primarily address domain shifts from scene-layout changes or sensor configurations. Cross-view LiDAR introduces viewpoint-driven geometric shifts that fundamentally alter the patterns of point-density variation and occlusion-induced missingness, beyond what current DG settings assume. Consequently, cross-view generalization remains a largely unexplored challenge.

%% file: sec/3_method.tex
\begin{figure*}[t]
    \centering
    \includegraphics[width=0.95\textwidth]{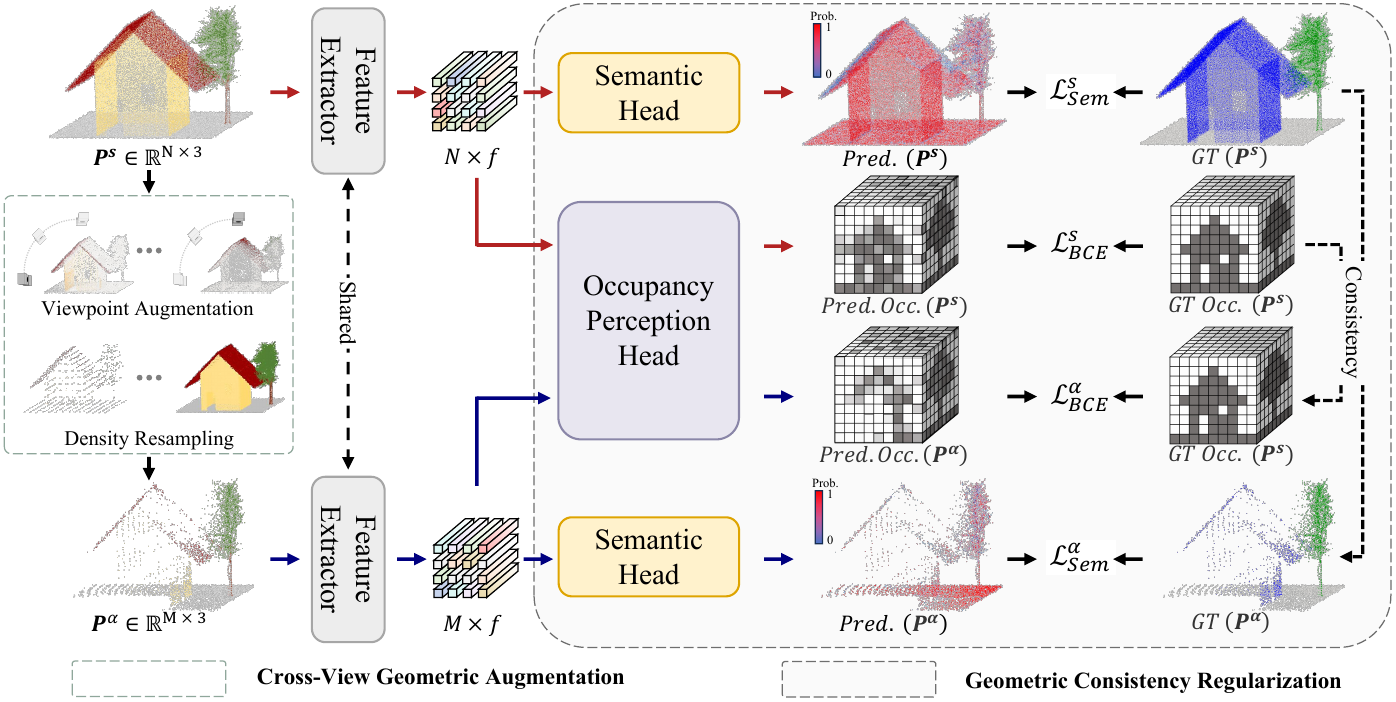}
    \caption{\textbf{Overview of the CVGC framework}. The framework consists of two main modules: (i) Cross-view Geometric Augmentation (CGA), which generates a view-dependent variant $\mathbf{P}^{\alpha}$ from the source point cloud $\mathbf{P}^s$; and (ii) Geometric Consistency Regularization (GCR), which enforces structural and semantic consistency across variants. A shared backbone predicts point-wise semantics for both views, with semantic consistency and voxel occupancy supervision jointly regularizing the learned representations, using only source-domain data.}
    \label{fig:stream}
\end{figure*}
\section{Method}
Cross-view LiDAR introduces viewpoint-dependent shifts in point density and structural missingness that existing DG methods cannot effectively handle. To address this challenge, we propose CVGC (Cross-View Geometric Consistency), a novel framework for cross-view domain generalization in LiDAR semantic segmentation. As shown in Fig.~\ref{fig:stream}, CVGC constructs geometric variations of the same scene from different viewpoints and enforces semantic and structural consistency across them to learn viewpoint-invariant representations that generalize to unseen point clouds from diverse viewpoints. In the following, we first define the cross-view DG problem for LSS in Section III-A, then detail the construction of view-dependent geometric variations in Section III-B, and finally present the corresponding consistency prediction strategy in Section III-C.

\subsection{Problem Formulation}
We consider a domain generalization setting for LiDAR semantic segmentation. 
The labeled source domain is defined as $ \mathcal{D}^s = \{(\mathbf{P}^s, \mathbf{Y}^s)\},$
where $\mathbf{P}^s$ denotes a source-domain point cloud scene and $\mathbf{Y}^s$ is its corresponding point-wise label set. The model is trained using only $\mathcal{D}^s$, without access to any target-domain data. The trained model is then evaluated on $K$ unseen and unlabeled target domains 
$\{\mathcal{D}_k^t\}_{k=1}^{K}$, where each $\mathcal{D}_k^t$ is drawn from a data distribution $p_k^t(\mathbf{P})$, while the source domain follows distribution $p^s(\mathbf{P})$, with $p_k^t(\mathbf{P}) \neq p^s(\mathbf{P})$. 
In cross-view LiDAR scenarios, the distribution shift arises not only from unseen scene layouts and sensor configurations, but is further amplified by differences in sensing viewpoints and acquisition geometries, which lead to distinct patterns of point-density variation and occlusion-induced missingness.

The objective of domain generalization for LiDAR semantic segmentation is to learn a segmentation function $f_\theta: \mathbb{R}^{M \times 3} \rightarrow \mathcal{C}^M$, parameterized by $\theta$, where $M$ is the number of points in a target scene and $\mathcal{C}$ denotes the label set. The model is trained by minimizing the empirical risk on the labeled source domain $\mathcal{D}^s$,
\begin{equation}
\min_{\theta} \; \mathbb{E}_{(\mathbf{P},\mathbf{Y})\sim \mathcal{D}^s}
\big[\mathcal{L}(f_\theta(\mathbf{P}), \mathbf{Y})\big],
\end{equation}
with the goal of achieving low expected risk on all unseen target domains. $\mathcal{L}(\cdot)$ is a point-wise segmentation loss function. Formally, the desired objective is
\begin{equation}
\mathbb{E}_{k}\Big[\, \mathbb{E}_{(\mathbf{P},\mathbf{Y}) \sim \mathcal{D}_k^t}
\big[\mathcal{L}(f_\theta(\mathbf{P}), \mathbf{Y})\big] \,\Big],
\end{equation}
Since no target-domain data are accessible during training, the learned parameters $\theta$ must induce representations that are robust to both unseen scene variations and cross-view geometric shifts, thereby enabling reliable generalization across heterogeneous sensing geometries.

\subsection{Cross-view Geometric Augmentation}

\begin{figure}[t]
    \centering
    \includegraphics[width=\columnwidth]{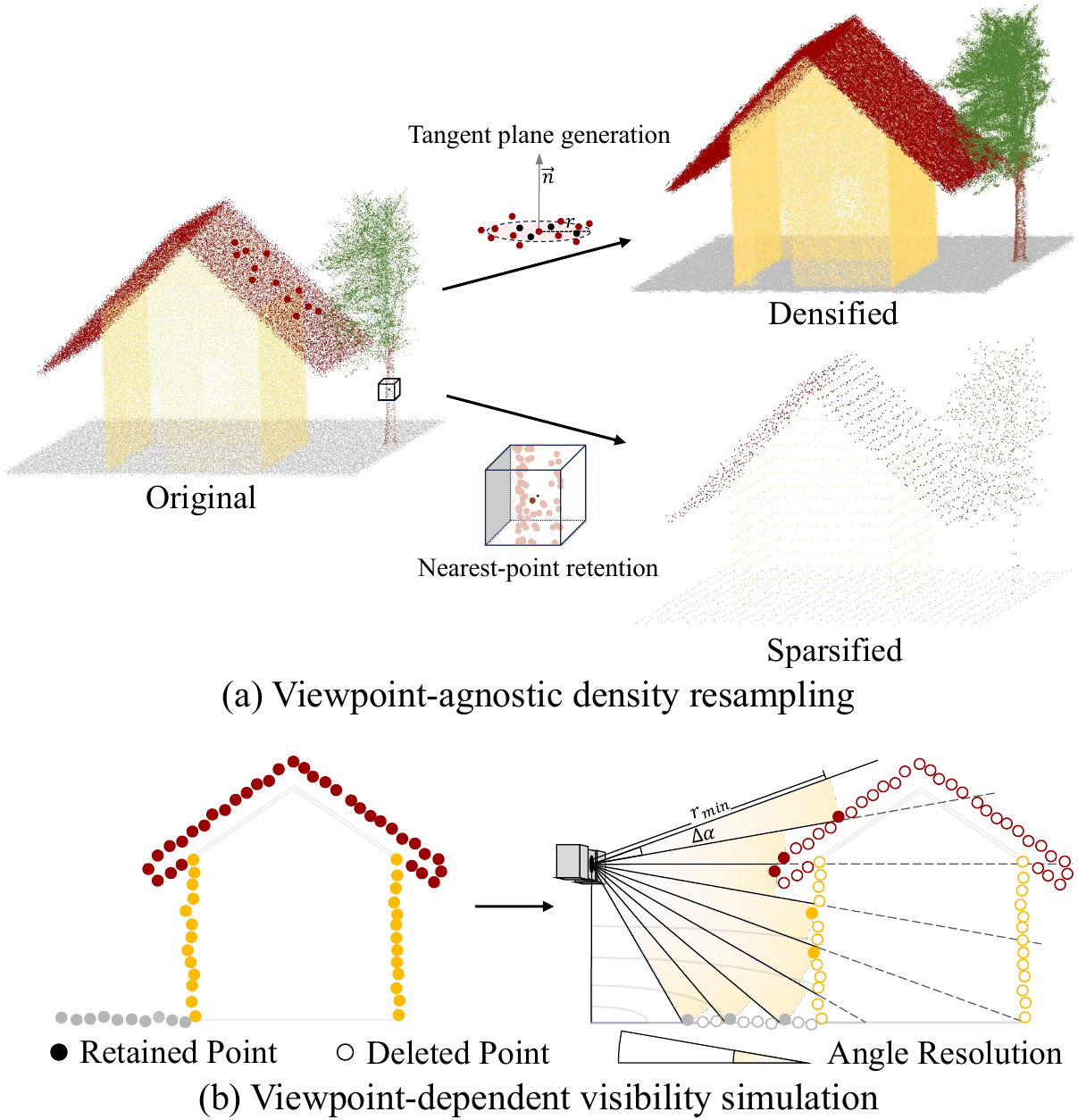}
    \caption{\textbf{Illustration of the CGA.} (a) Viewpoint-agnostic density resampling, which adjusts point cloud scene density via tangent plane-based densification and voxel-based sparsification. (b) Viewpoint-dependent visibility simulation, which constructs occluded scenes based on a spherical projection process.}
    \label{fig:simulate}
\end{figure}
The goal of this module is to generate view-dependent geometric variants of source-domain point clouds, allowing the model to observe the same scene under diverse sensing viewpoints during training. Existing LiDAR augmentation methods are largely platform-specific, primarily designed for vehicle-mounted LiDAR and fixed scanning geometries. In cross-view settings, heterogeneous platforms produce fundamentally different density distributions and coverage patterns, rendering current simulation strategies ineffective. To overcome this limitation, we propose \textbf{Cross-view Geometric Augmentation (CGA)}, which integrates (i) viewpoint-agnostic density resampling via geometry-aware densification and sparsification, and (ii) viewpoint-dependent visibility simulation to model occlusion and structural missingness, thereby enhancing cross-view training diversity and reducing overfitting to specific sensing viewpoints.

\subsubsection{Viewpoint-agnostic density resampling}
Point density plays a critical role in LiDAR semantic segmentation, yet existing methods typically assume that density variations follow similar patterns across domains. For example, vehicle-mounted LiDAR usually exhibits radial decay—dense near the sensor and sparse at long range—whereas airborne LiDAR tends to produce relatively uniform density on planar surfaces. In cross-view scenarios, such platform-dependent assumptions break down, as density distributions are governed by fundamentally different sensing geometries and visibility conditions. Consequently, current density modeling strategies tailored to specific acquisition settings cannot generalize effectively to heterogeneous viewpoints. To address this issue, we propose a viewpoint-agnostic density augmentation strategy composed of two operations: Densification and Sparsification. Together, they produce diverse density variants that preserve scene geometry and enhance robustness to cross-view density shifts.

\textbf{Densification.}
For a given source-domain scene $\mathbf{P}_i^s$, as illustrated in the upper part of Fig.~\ref{fig:simulate}(a), we perform density upsampling by locally generating additional points on the estimated tangent plane at each point.
Concretely, for every point $\mathbf{x}_{ij} \in \mathbf{P}_i^s$, a surface normal $\mathbf{n}_{ij}$ is estimated using a KNN-based local covariance (PCA) method, and an orthonormal local frame $\{\mathbf{u}_{ij}, \mathbf{v}_{ij}, \mathbf{n}_{ij}\}$ is constructed, where $\mathbf{u}_{ij} \perp \mathbf{n}_{ij}$ and $\mathbf{v}_{ij} = \mathbf{n}_{ij} \times \mathbf{u}_{ij}$.
We then generate $K$ new samples around $\mathbf{x}_{ij}$ by drawing independent random variables $\xi_{ij}^{(1)}, \xi_{ij}^{(2)} \sim \mathcal{U}(0,1)$ and defining a polar perturbation on the tangent plane with radius $\rho_{ij} = r \sqrt{\xi_{ij}^{(1)}}$ and angle $\theta_{ij} = 2\pi \xi_{ij}^{(2)}$, where $r$ controls the maximum disturbance magnitude. Each synthesized point is given by
\begin{equation}
\tilde{\mathbf{x}}_{ij}^{(k)} = \mathbf{x}_{ij} + \rho_{ij}^{(k)} \Bigl( \cos\theta_{ij}^{(k)} \cdot \mathbf{u}_{ij} + \sin\theta_{ij}^{(k)} \cdot \mathbf{v}_{ij} \Bigr),
\end{equation}
This ensures an approximately uniform distribution within a disk of radius \(r\) on the local tangent plane while preserving the underlying surface geometry.
Collecting all generated samples across all original points yields an augmented point set 
$
\mathbf{P}_i^\alpha = \mathbf{P}_i^s \ \cup\ \{ \tilde{\mathbf{x}}_{ij}^{(k)} \ |\  j=1,\dots,M_i,\ k=1,\dots,K \}.
$

\textbf{Sparsification.}
For a given source-domain scene $\mathbf{P}_i^s$, as illustrated in the lower part of Fig.~\ref{fig:simulate}(a), we perform density downsampling by retaining, within each voxel of size 
$v$, the point that is nearest to the voxel centroid.
Concretely, Each point $\mathbf{x}_{ij} \in \mathbf{P}_i^s$ is assigned a voxel index $\mathbf{u}_{ij} = \left\lfloor \mathbf{x}_{ij} / v \right\rfloor \in \mathbb{Z}^3$. Within each non-empty voxel $\mathbf{u}$, we first compute the centroid of the points belonging to the current point cloud $\mathbf{P}_i^s$:
\begin{equation}
    \mathbf{c}(\mathbf{u}) = \frac{1}{|\mathcal{P}(\mathbf{u})|} \sum_{j \in \mathcal{P}(\mathbf{u})} \mathbf{x}_{ij},
\end{equation}
where \(\mathcal{P}(\mathbf{u}) = \{j \mid \mathbf{u}_{ij} = \mathbf{u}\}\) is the set of point indices in voxel $\mathbf{u}$.
We retain the point closest to the centroid:
\begin{equation}
\mathbf{x}^*(\mathbf{u}) = \mathbf{x}_{i j^*}, 
\quad j^* = \arg\min_{j \in \mathcal{P}(\mathbf{u})} \|\mathbf{x}_{ij} - \mathbf{c}(\mathbf{u})\|_2.
\end{equation}
The down-sampled point cloud for the $i$-th scene is
$
\mathbf{P}_{i}^{\alpha} = \bigl\{ \mathbf{x}^*(\mathbf{u}) \mid \mathcal{P}(\mathbf{u}) \neq \emptyset \bigr\}.
$
This process ensures exactly one representative point per occupied voxel while preserving local geometric structure.

\subsubsection{Viewpoint-dependent visibility simulation}
Density resampling alone cannot capture structural changes caused by viewpoint shifts. In cross-view scenarios, different platforms observe scenes from fundamentally different perspectives, leading to drastic variations in geometric visibility and coverage. As a result, many structures visible in one viewpoint may be partially or entirely missing in another, producing occlusion-induced missingness that cannot be addressed by density adjustment alone. Existing augmentation methods largely assume similar viewpoints and therefore lack mechanisms to simulate such viewpoint-dependent structural differences. Consequently, they are unable to model the distinct visibility patterns required for effective cross-view generalization. To address this limitation, we introduce a viewpoint-dependent visibility simulation scheme based on spherical projection. The proposed method is platform-agnostic and simulates generic viewpoint-driven visibility and occlusion, producing point cloud variants that capture structural incompleteness across diverse sensing viewpoints.

For a given source-domain scene $\mathbf{P}_i^s$, we first randomly select a ground point in the scene as the planar reference and then sample a height value to construct a virtual viewpoint $\mathbf{v} \in \mathbb{R}^3$. The entire scene is subsequently transformed into a viewpoint-centered coordinate system by translating each point as $\tilde{\mathbf{x}}_{ij}=\mathbf{x}_{ij}-\mathbf{v}$. Next, $\tilde{\mathbf{x}}_{ij}$ is converted into spherical coordinates $(r_{ij},\theta_{ij},\phi_{ij})$, where
\begin{equation}
\left\{
\begin{aligned}
    r_{ij} &= |\tilde{\mathbf{x}}_{ij}|_2, \\
    \theta_{ij} &= \arccos(\tilde{z}_{ij}/r_{ij}) \in [0,\pi], \\
    \phi_{ij} &= \mathrm{atan2}(\tilde{y}_{ij},\tilde{x}_{ij}) \in [0,2\pi).
\end{aligned}
\right.
\end{equation}

The angular space is then discretized with resolution $\Delta\alpha$ by computing the bin indices $\hat{\theta}_{ij}=\lfloor \theta_{ij}/\Delta\alpha \rfloor, \hat{\phi}_{ij}=\lfloor \phi_{ij}/\Delta\alpha \rfloor$. Each discretized angular bin $(\hat{\theta},\hat{\phi})$ is assigned a unique index $h=\hat{\theta}M_\phi+\hat{\phi}$, where $M_\phi=\lceil 2\pi/\Delta\alpha\rceil$. Finally, the visible point set is obtained by retaining, for each non-empty angular bin, only the point with the minimum radial distance $r_{ij}$ along that viewing ray:
\begin{equation}
    \mathbf{P}_i^\alpha = \bigl\{ (i,j) \mid r_{ij} = \min \{ r_{k\ell} \mid h_{k\ell} = h_{ij} \} \bigr\},
\end{equation}
where $h_{ij} = \hat{\theta}_{ij} \cdot M_\phi + \hat{\phi}_{ij}$. This process effectively simulates viewpoint-dependent visibility by keeping only the closest surface point along each discretized viewing ray.

\subsection{Geometric Consistency Regularization}
\begin{figure*}[t]
    \centering
    \includegraphics[width=0.9\textwidth]{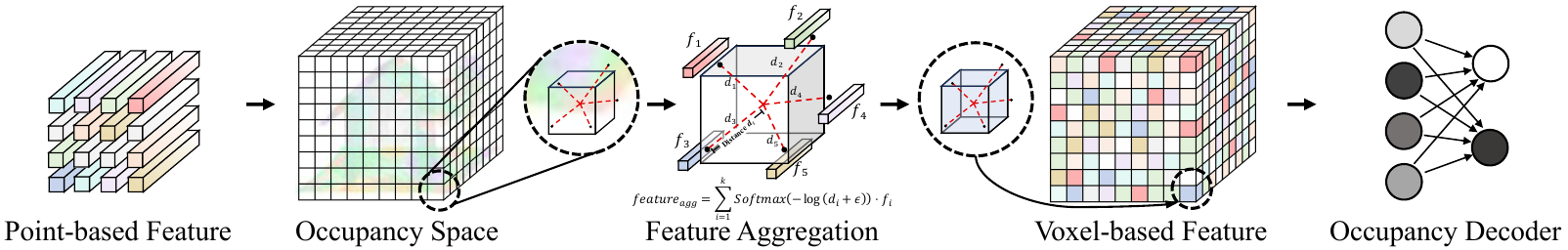}
    \caption{\textbf{Schematic of the occupancy perception head.} Point-wise features are aggregated into voxel representations via voxelization and KNN-based interpolation, and processed by sparse convolutional layers to predict voxel occupancy, providing unified geometric supervision across different viewpoints.}
    \label{fig:Occ}
\end{figure*}

Cross-view geometric augmentation produces multiple variants of the same scene under different viewpoints. Since these variants originate from identical underlying geometry, their structural and semantic predictions should remain consistent. However, direct alignment of local features across views is difficult due to viewpoint-dependent occlusions and sampling patterns. To exploit this inherent correspondence, we propose \textbf{Geometric Consistency Regularization (GCR)}, which enforces consistency among cross-view variants at both the structural and semantic levels. Specifically, features from different viewpoints are first fed into an occupancy perception head (Fig.~\ref{fig:Occ}), which performs Occupancy Space Construction and Voxel Feature Aggregation to produce voxel-level representations, and predicts voxel-wise occupancy states. Based on these occupancy predictions, Consistency Supervision enforces structural consistency across view variants, while semantic consistency is applied directly to point-wise segmentation outputs. By jointly regularizing voxel-level structure and point-level semantics for different variants of the same scene, the model learns viewpoint-invariant representations and generalizes robustly to unseen sensing viewpoints.

\subsubsection{Occupancy Space Construction}
Given the source-domain point cloud $\mathbf{P}^s_i = \{\mathbf{x}_{ij} \in \mathbb{R}^3\}_{j=1}^{M_i}$, we discretize the bounded scene into a cubic voxel grid with resolution $v$. We define the integer voxel coordinate for each point $\mathbf{x}_{ij}$ as $\mathbf{u}_{ij} = \left\lfloor \mathbf{x}_{ij} / v \right\rfloor \in \mathbb{Z}^3$. Based on this, an occupancy space $\mathcal{V} = \{\mathbf{u} \mid \mathbf{u} \in \Omega \subset \mathbb{Z}^3\}$ is constructed over a restricted spatial extent $\Omega \subset \mathbb{Z}^3$. Each voxel $\mathbf{u} \in \Omega$ is assigned a binary occupancy label by the indicator function
\begin{equation}
    \mathcal{O}(\mathbf{u}) = \mathbb{I}\left( \exists \, \mathbf{x}_{ij} \in \mathbf{P}^s_i \ \text{s.t.} \ \left\lfloor \mathbf{x}_{ij} / v \right\rfloor = \mathbf{u} \right) \in \{0,1\},
\end{equation}
where $\mathcal{O}(\mathbf{u})=1$ denotes occupied space and $\mathcal{O}(\mathbf{u})=0$ denotes empty but observable free space within $\Omega$, yielding a self-supervised geometric ground truth that explicitly encodes source-domain visibility and occlusion structure.

\subsubsection{Voxel Feature Aggregation}
      To provide dense geometric representations for all voxels in the occupancy space $\mathcal{V}$, including both occupied and unoccupied regions, we aggregate point-wise features extracted from the backbone network applied to the input point cloud $\mathbf{P}$ (either the source $\mathbf{P}^s_i$ or its augmented $\mathbf{P}^a_i$) via distance-weighted $k$-nearest neighbor interpolation. Let $\mathbf{F} \in \mathbb{R}^{M \times D}$ denote the per-point features, where $M$ is the number of points and $D$ is the feature dimension. For each voxel $\mathbf{u}\in\mathcal{V}$ produced by voxelization we retrieve its $k$ nearest points $\{\mathbf{x}_{(l)},\mathbf{f}_{(l)}\}_{l=1}^k$ from $\mathbf{P}$ based on Euclidean distance $d_{(l)}$ in the discretized coordinate space, compute inverse-distance weights $w_{(l)}=(d_{(l)}+\epsilon)^{-1}$ and their normalized form 
\begin{equation}
\tilde{w}_{(l)}=\frac{w_{(l)}}{\sum_{m=1}^k w_{(m)}},
\end{equation}
The aggregated voxel feature is then obtained by distance-weighted averaging:
\begin{equation}
\mathbf{f}_{\mathbf{u}} = \sum_{l=1}^k \tilde{w}_{(l)} \mathbf{f}_{(l)}.
\end{equation}
This aggregation is performed uniformly across all voxels in \(\mathcal{V}\), ensuring that even unoccupied voxels (where \(\mathcal{O}(\mathbf{u})=0\)) receive interpolated semantic features from nearby points. The resulting voxel features \(\{\mathbf{f}_{\mathbf{u}}\}\), together with their corresponding voxel coordinates \(\{\mathbf{u}\}\), are formed into a sparse tensor and fed into a sparse convolutional network to predict voxel occupancy.

\subsubsection{Consistency Supervision}
To capture local geometric information, we employ a shared sparse convolutional network to process voxel features extracted from the source point cloud $\mathbf{P}^s_i$ and its geometrically enhanced version $\mathbf{P}^\alpha_i$. The network consists of two $1\times1\times1$ sparse convolutional layers, each followed by batch normalization and ReLU activation, and produces a voxel-wise occupancy prediction $\hat{\mathcal{O}}(\mathbf{u})$ for each voxel $\mathbf{u} \in \mathcal{V}$. 
The occupancy prediction is supervised using a binary cross-entropy (BCE) loss, where the supervision signal is provided by a self-supervised occupancy ground truth $\mathcal{O}(\mathbf{u})$, constructed solely from the source point cloud. The BCE loss is defined as:
\begin{equation}
\begin{split}
\mathcal{L}_{\text{BCE}}(\hat{\mathcal{O}}, \mathcal{O}) &= -\sum_{\mathbf{u} \in \mathcal{V}} \Big[ \mathcal{O}(\mathbf{u}) \log \hat{p}_1(\mathbf{u}) \\
&\quad + (1 - \mathcal{O}(\mathbf{u})) \log (1 - \hat{p}_1(\mathbf{u})) \Big],
\end{split}
\label{eq:loss_BCE}
\end{equation}
where $\hat{p}_1(\mathbf{u}) = \hat{\mathcal{O}}(\mathbf{u})$ denotes the predicted probability of voxel $\mathbf{u}$ being occupied.

In addition to occupancy supervision, we apply semantic supervision $\mathcal{L}_{\text{sem}}$ to both the source view and its augmented variants. While semantic labels provide category-level discrimination, occupancy supervision offers complementary geometry-aware guidance that is robust to semantic ambiguity. By jointly enforcing semantic and occupancy consistency across geometric augmentations, the network learns representations that are invariant to viewpoint-induced distortions. Finally, the overall loss for CVGC is defined as:
\begin{equation}
\mathcal{L}_{\text{total}}=\mathcal{L}_{\text{sem}}^{s}+\mathcal{L}_{\text{sem}}^{\alpha}+\mathcal{L}_{\text{BCE}}^s +\mathcal{L}_{\text{BCE}}^\alpha,
\label{eq:loss_overall}
\end{equation}
where $\mathcal{L}_{\text{BCE}}^s$ = $ \mathcal{L}_{\text{BCE}}(\hat{\mathcal{O}}^s, \mathcal{O})$ and $\mathcal{L}_{\text{BCE}}^{\alpha}$ = $ \mathcal{L}_{\text{BCE}}(\hat{\mathcal{O}}^{\alpha}, \mathcal{O})$. This joint optimization promotes consistent predictions across views and improves cross-view generalization.

%% file: sec/4_experiments.tex
\section{Experiments}
\subsection{Datasets}

\begin{figure}[t]
    \centering
    \includegraphics[width=0.5\textwidth]{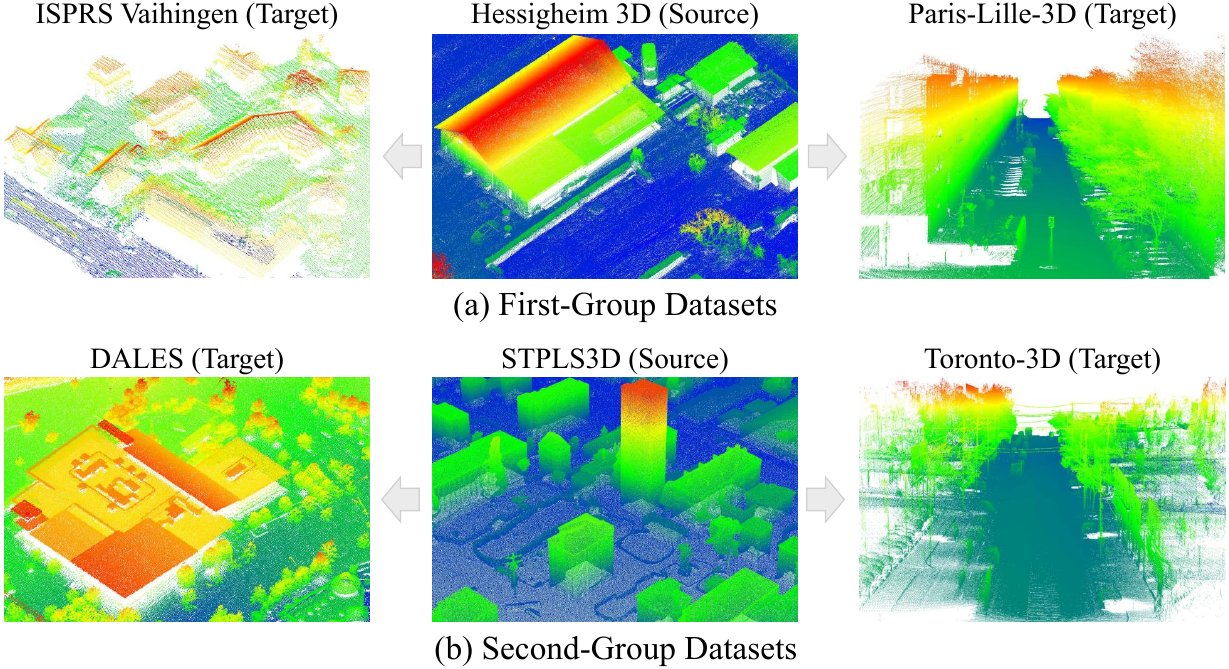}
    \caption{\textbf{Benchmark for cross-view domain generalization.} The six datasets are partitioned into two groups following a one-source, two-target protocol. (a) First-Group Datasets: H3D (UAV) \(\rightarrow\) {Paris-Lille-3D (MLS) and ISPRS Vaihingen (ALS)}. (b) Second-Group Experiments: STPLS3D (synthetic UAV) \(\rightarrow\) {Toronto-3D (MLS) and DALES (ALS)}.}
    \label{fig:dataset}
\end{figure}

Existing DG studies for LiDAR semantic segmentation are typically evaluated on datasets with similar sensing viewpoints and usually adopt a single-source, single-target protocol~\cite{xiao2022polarmix,zhao2024unimix,kim2023single}. Such settings are inadequate for cross-view domain generalization, where target domains differ fundamentally in acquisition geometry. To address this gap, we construct, to the best of our knowledge, the first benchmark specifically designed for cross-view DG, comprising six LiDAR datasets from heterogeneous platforms, including airborne, vehicle-mounted, and UAV-based systems. The datasets are divided into two groups, each adopting a one-source, two-target protocol, where all three datasets within a group originate from different scene layouts and sensing viewpoints. This setup evaluates the ability of models to generalize to multiple unseen domains with fundamentally different acquisition views.

The first group comprises Hessigheim 3D (H3D)~\cite{kolle2021hessigheim}, Paris-Lille-3D~\cite{roynard2018paris} and ISPRS Vaihingen~\cite{rottensteiner2012isprs}, which correspond to acquisition perspectives of UAV-, vehicle- and airborne-mounted LiDAR, respectively, as shown in Fig.~\ref{fig:dataset} (a). \textbf{H3D}, captured with a UAV-mounted LiDAR system, is a rural-scene dataset with relatively high point density ($\approx$ 800 pts/m$^{2}$) and 11 semantic classes. \textbf{Paris-Lille-3D} is a typical vehicle-mounted LiDAR urban dataset with an average point density exceeding 1000 pts/m$^{2}$ and up to 50 annotated semantic classes. \textbf{ISPRS Vaihingen} is an airborne LiDAR dataset with relatively low but uniform point density (4–7 pts/m$^{2}$) and 9 semantic classes. In this group, H3D is set as the source domain, with Paris-Lille-3D and ISPRS Vaihingen serving as target domains. For cross-dataset evaluation the original semantic labels of the three datasets are remapped to five shared classes: \textit{Ground, Building, Natural, Vehicle, and Urban Furniture}.

The second group consists of the STPLS3D synthetic subset~\cite{chen2022stpls3d}, Toronto-3D~\cite{Tan2020toronto} and DALES~\cite{varney2020dales}, as shown in Fig.~\ref{fig:dataset} (b). The \textbf{STPLS3D} synthetic subset is generated by procedural modeling and simulated UAV flight paths, comprising 18 semantic classes. \textbf{Toronto-3D} is a vehicle-mounted urban LiDAR dataset with eight semantic classes and an average point density of approximately 1000 pts/m$^{2}$. \textbf{DALES} is an airborne LiDAR dataset with eight semantic classes and an average point density of about 50 pts/m$^{2}$. In this group the STPLS3D synthetic subset (V1–V3) is chosen as the source domain, while Toronto-3D and DALES are used as target domains. Beyond cross-view discrepancies, this setup introduces a substantially larger domain gap, as the source data are synthetic whereas the targets are real-world datasets. For cross-dataset evaluation labels are remapped to six shared classes: \textit{Ground, Building, Natural, Vehicle, Pole, and Fence}.

\subsection{Implementation Details}
All models are implemented in the PyTorch framework and executed on an NVIDIA GeForce RTX 4090 GPU with 24 GB memory. For density resampling, we randomly sample an average inter-point distance within the range of $[0.01, 0.5]$ m to simulate diverse point density distributions. For viewpoint-dependent visibility simulation, view heights are randomly sampled from ${2.0, 4.0, 8.0, 16.0, 32.0, 64.0}$ m, and occlusions are simulated using an angular resolution of 0.01 rad to emulate structurally incomplete point clouds. MinkUNet34~\cite{choy20194d} is adopted as the backbone for point cloud semantic feature extraction. The occupancy prediction head consists of a two-layer sparse convolutional structure. The voxel size is set to 1.0 m, and feature aggregation is performed using KNN with K=3. For dataset-specific settings, the first group uses a patch size of $32$ m $\times 32$ m with a 50\% overlap between training blocks and a voxel size of 0.25 m. The second group adopts a larger patch size of $50$ m $\times 50$ m and a voxel size of 0.3 m. During training, a fixed learning rate of $1 \times 10^{-3}$ is used with the Adam optimizer, and the batch size is set to 4.

\subsection{Comparative Methods}
We compare the proposed method with five representative approaches for LiDAR semantic segmentation in the cross-view DG setting, including one source-only baseline, one domain adaptation (DA) method, and three domain generalization (DG) methods. The source-only baseline is trained solely on labeled source-domain data and directly evaluated on unseen targets. For DA, we adopt CosMix~\cite{saltori2022cosmix}, which mitigates domain gaps by exchanging scene segments between labeled source and unlabeled target data. For DG, we evaluate PolarMix~\cite{xiao2022polarmix}, which performs circular mixing along the azimuth direction to simulate LiDAR scanning patterns; DGLSS~\cite{kim2023single}, which employs sparsity-driven augmentation and semantic consistency constraints; and UniMix~\cite{zhao2024unimix}, which constructs intermediate bridge domains for learning domain-invariant representations. To ensure a fair comparison, all methods use the same backbone and are trained with only point coordinates as input.

For quantitative evaluation, we report per-class Intersection over Union (IoU) and mean IoU (mIoU). The per-class IoU is defined as
\begin{equation}
\text{IoU}_c = \frac{\text{TP}_c}{\text{TP}_c + \text{FP}_c + \text{FN}_c},
\end{equation}
where $\text{TP}_c$, $\text{FP}_c$, and $\text{FN}_c$ denote the numbers of true positives, false positives, and false negatives for class $c$, respectively. The mean IoU (mIoU) is computed by averaging $\text{IoU}_c$ over all classes.

\subsection{Results and Analysis}
\begin{figure*}[t]
    \centering
    \includegraphics[width=\textwidth]{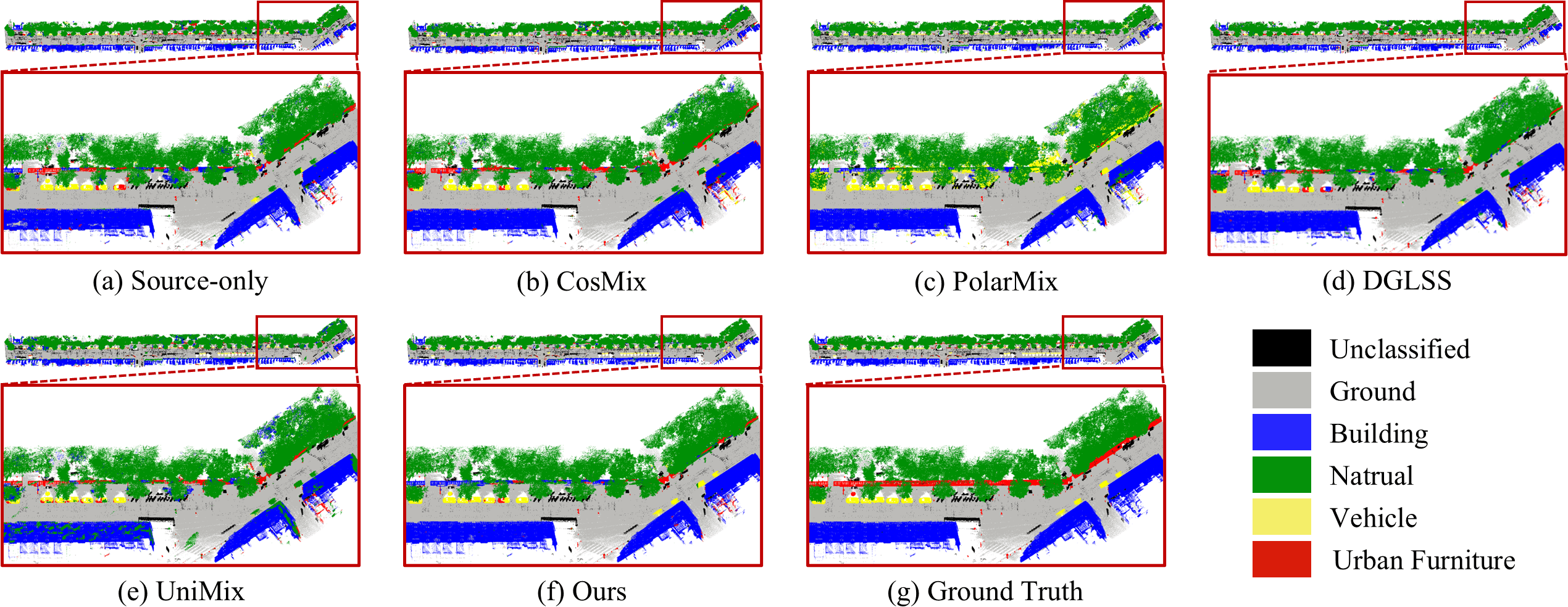}
    \caption{\textbf{Qualitative results for Hessigheim-3D → Paris-Lille-3D.} Results of different methods are visualized on the MLS target domain, where colors indicate semantic categories.}
    \label{fig:PL3D}
\end{figure*}

\input{table/Method_H3D}

\begin{figure*}[t]
    \centering
    \includegraphics[width=\textwidth]{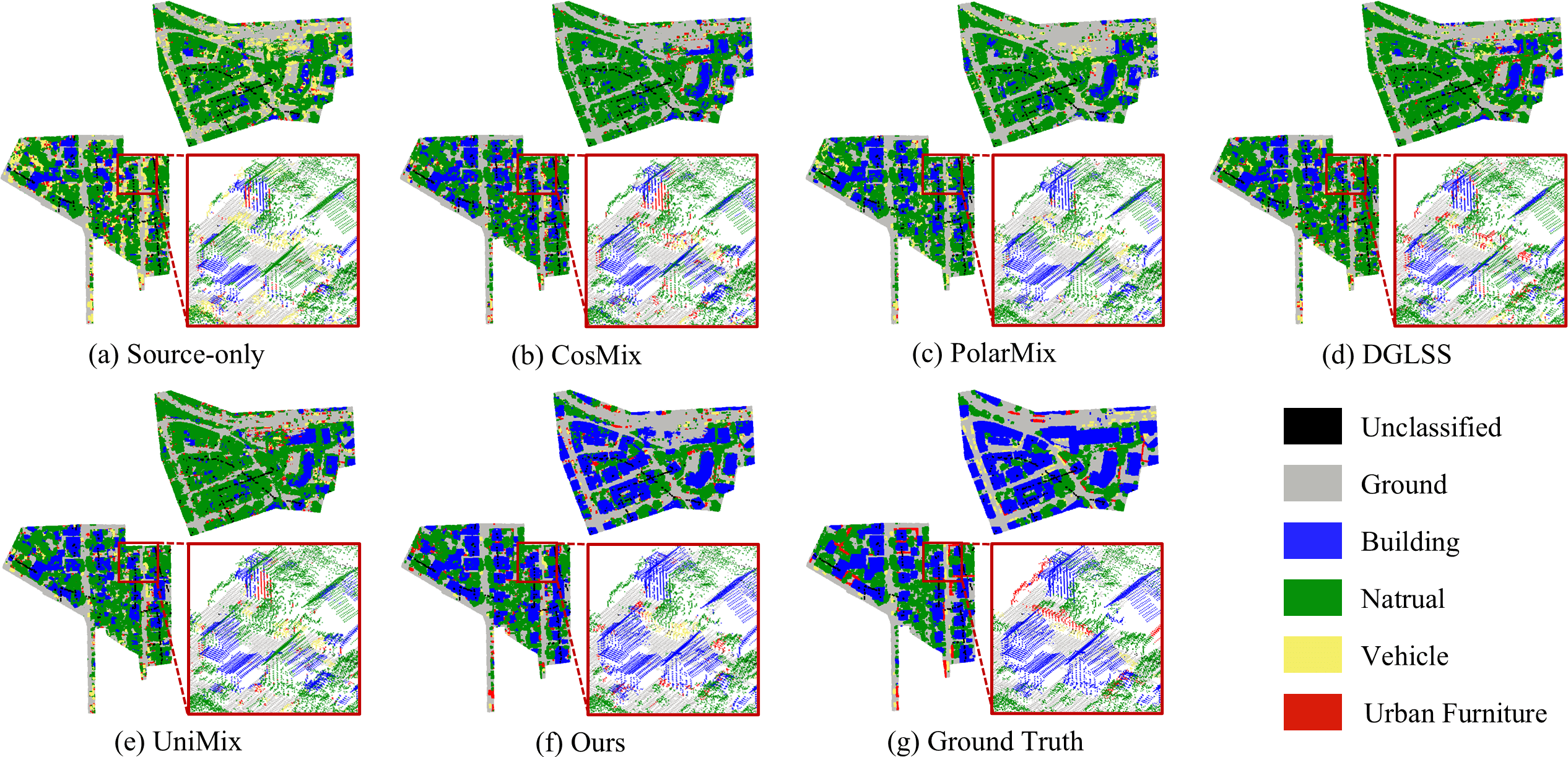}
    \caption{\textbf{Qualitative results for Hessigheim-3D → ISPRS Vaihingen.} Results of different methods are visualized on the ALS target domain, where colors indicate semantic categories.}
    \label{fig:ISPRS}
\end{figure*}

\subsubsection{First-Group Experiments}
When the model trained on the source domain is directly applied to the target domains, substantial performance discrepancies are observed, as shown in Fig.~\ref{fig:PL3D}(a) and Fig.~\ref{fig:ISPRS}(a). As reported in Table~\ref{tab:h3d_results}, the source-only method achieves 66.06\% mIoU on Paris-Lille-3D, while its performance drops sharply to 31.20\% mIoU on ISPRS Vaihingen. This indicates that the performance degradation strongly depends on the extent of cross-view discrepancy. Since mIoU assigns equal importance to all categories, it is particularly sensitive to classes with complex geometries or limited spatial support, which are more susceptible to view-induced structural variations.

Building upon this baseline, the proposed approach improves segmentation performance on both target domains, as illustrated in Fig.~\ref{fig:PL3D}(f) and Fig.~\ref{fig:ISPRS}(f). On Paris-Lille-3D, it achieves 72.87\% mIoU, corresponding to an improvement of 6.81\%. On ISPRS Vaihingen, where the point density differs from the source domain by nearly two orders of magnitude, the performance gain is substantially larger, reaching 55.18\% mIoU, i.e., an improvement of 23.98\%. This observation suggests that the proposed method is particularly effective in mitigating severe view-induced distribution shifts rather than merely refining already aligned cases.

As shown in Fig.~\ref{fig:PL3D}(f) and Fig.~\ref{fig:ISPRS}(f), performance gains are observed across most semantic categories. Notably, the Vehicle class exhibits consistent IoU improvements exceeding 20\% on both datasets, indicating enhanced object-level generalization under viewpoint changes. On ISPRS Vaihingen, the Building category achieves a remarkable 54.06\% IoU increase, while the Natural category improves by 31.49\%, reflecting the method’s ability to preserve geometric cues under extreme density discrepancies. In contrast, gains on the Urban Furniture class remain limited, which can be attributed to inconsistencies in subcategory definitions across datasets rather than geometric misalignment, as visually evidenced by the red regions in Fig.~\ref{fig:PL3D}(g) and Fig.~\ref{fig:ISPRS}(g).

Existing DG methods assume similar acquisition settings and thus perform unstably under large viewpoint shifts. They show moderate improvements on ISPRS Vaihingen, which shares a viewpoint similar to the source, but suffer clear degradation on Paris-Lille-3D with substantially different sensing geometry. Among them, PolarMix attains the largest gain on ISPRS Vaihingen (6.14\% mIoU, Fig.~\ref{fig:ISPRS}(c)) but experiences a severe drop on Paris-Lille-3D (25.66\% mIoU, Fig.~\ref{fig:PL3D}(c)), indicating limited robustness to drastic cross-view shifts. UniMix and DGLSS attempt to enhance robustness by simulating cross-domain variations at the data level, leading to more stable performance compared with aggressive mixing strategies. However, their overall improvements remain limited, suggesting that source-domain simulation and data mixing alone are insufficient to fully compensate for the structural discrepancies induced by significant viewpoint changes.

In contrast, the proposed method demonstrates consistent improvements across both moderate and extreme cross-view scenarios without relying on unlabeled target data. While the domain adaptation method CoSMix benefits from access to target data and achieves notable gains on Paris-Lille-3D, it relaxes deployment assumptions. Overall, these results indicate that explicitly preserving geometric continuity and structural plausibility across viewpoints is critical for effective cross-view LiDAR semantic segmentation, and that the proposed approach provides a more robust solution under diverse acquisition conditions.

\subsubsection{Second-Group Experiments}
\begin{figure*}[t]
    \centering
    \includegraphics[width=\textwidth]{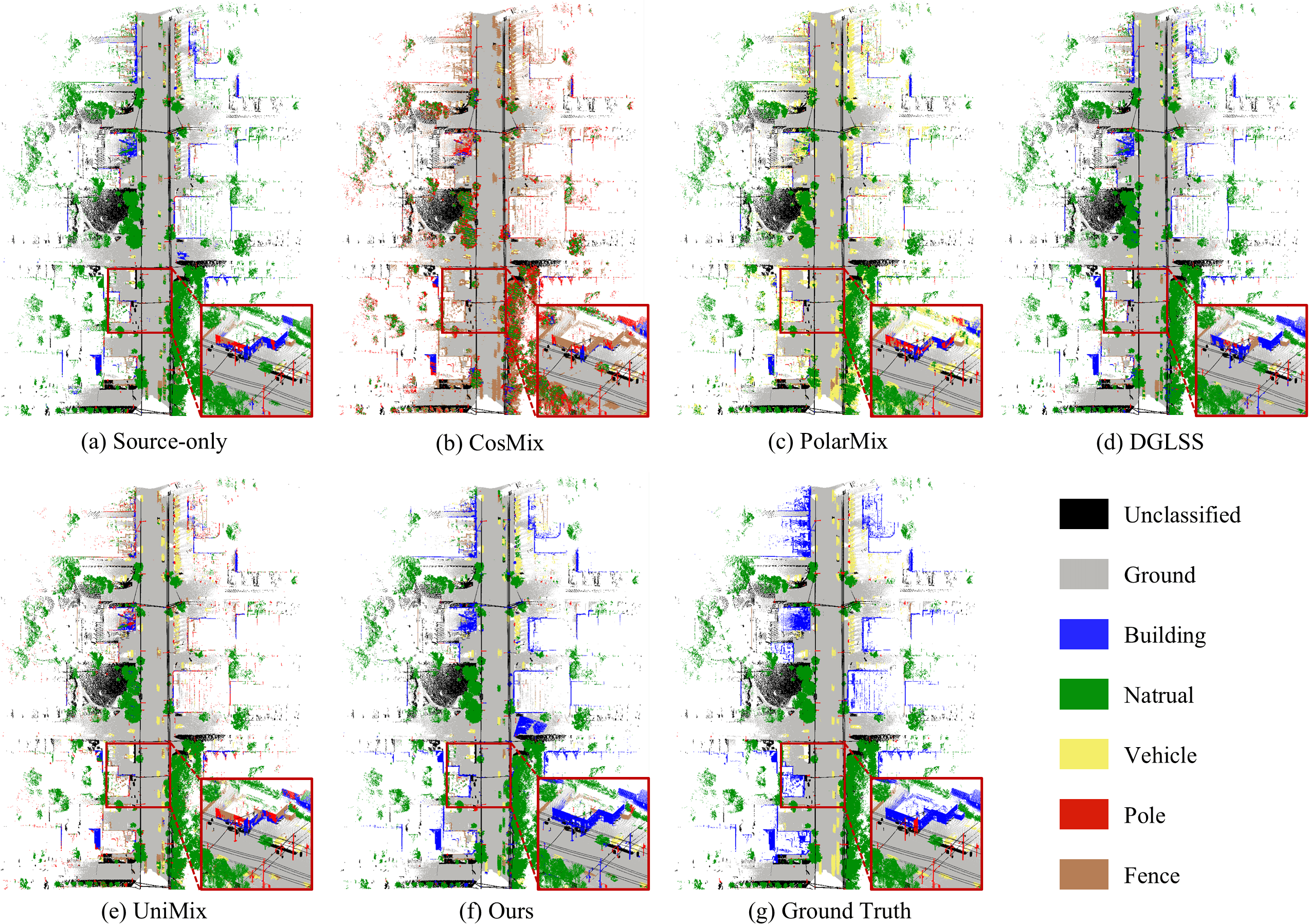}
    \caption{\textbf{Qualitative results for STPLS3D → Toronto-3D.} Results of different methods are visualized on the MLS target domain, where colors indicate semantic categories.}
    \label{fig:T3D}
\end{figure*}

\begin{figure*}[t]
    \centering
    \includegraphics[width=\textwidth]{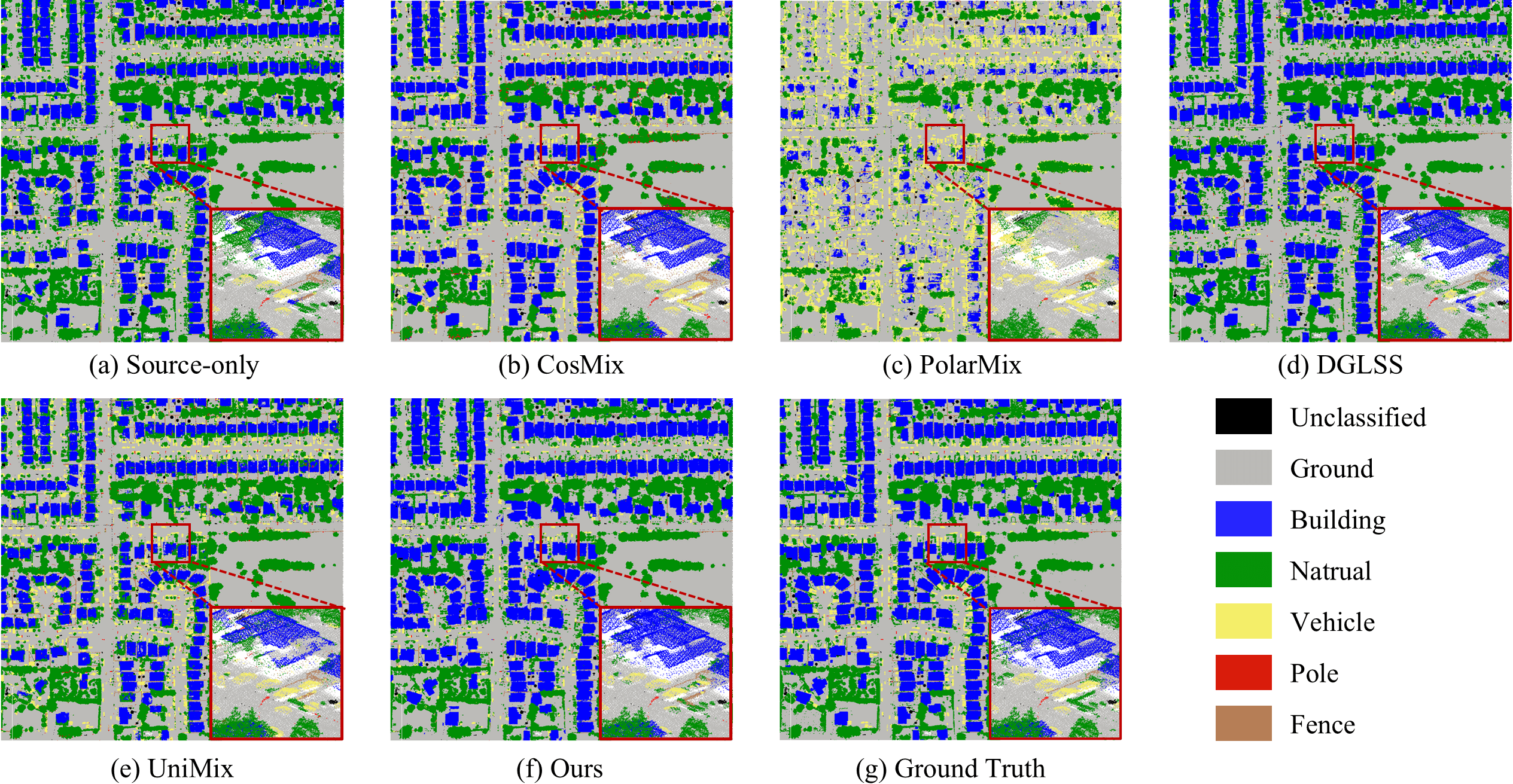}
    \caption{\textbf{Qualitative results for STPLS3D → DALES.} Results of different methods are visualized on the ALS target domain, where colors indicate semantic categories.}
    \label{fig:DALES}
\end{figure*}
\input{table/Method_STPLS3D}

As illustrated in Fig.~\ref{fig:T3D}(a) and Fig.~\ref{fig:DALES}(a), directly applying the model trained on the synthetic source domain to real-world target domains leads to substantial performance discrepancies. As reported in Table~\ref{tab:stpls3d_results}, the baseline achieves 47.79\% mIoU on Toronto-3D and 58.57\% mIoU on DALES. Unlike Experiment A, where the domain gap is dominated by extreme point density differences, the performance discrepancies observed here mainly arise from fundamental structural inconsistencies between synthetic photogrammetric data and real-world acquisition modalities, particularly under differing viewpoints.

Building upon the baseline, the proposed approach achieves stable improvements on both real-world target datasets, reaching 51.28\% mIoU on Toronto-3D and 64.42\% mIoU on DALES. Notably, the magnitude of performance gains correlates with the degree of perspective alignment between the synthetic source domain and the target data. On DALES (Fig.~\ref{fig:DALES}), which shares a similar aerial acquisition perspective with the source domain, mIoU improves by 5.85\%, with consistent gains observed across all semantic categories. In contrast, Toronto-3D is collected from a near-ground viewpoint (Fig.~\ref{fig:T3D}), resulting in more pronounced facade–roof structural discrepancies; nevertheless, the proposed method still achieves an improvement of 3.49\% mIoU. These results indicate that the method effectively adapts to varying degrees of perspective-induced structural divergence, rather than being limited to well-aligned acquisition settings.

As shown in Fig.~\ref{fig:T3D}(f) and Fig.~\ref{fig:DALES}(f), performance gains are concentrated in categories characterized by stable and distinguishable geometric structures. In particular, the Vehicle category (yellow) exhibits consistent and substantial IoU improvements across both datasets, reflecting enhanced object-level generalization under viewpoint changes. The Building category (blue) also benefits significantly, achieving an IoU gain of 15.16\% on DALES and 9.52\% on Toronto-3D, indicating improved modeling of facade–roof structural relationships. Additionally, moderate improvements are observed for the Fence category (brown) on Toronto-3D (IoU +4.72\%), where viewpoint-dependent structural exposure plays a critical role. In contrast, the Pole category (red) remains particularly challenging due to severe occlusions and missing structures caused by perspective variations, resulting in limited and uneven improvements. Consistent with the first group of experiments, underscoring the method’s effectiveness in enhancing performance on minority and structurally complex categories that are most sensitive to viewpoint-induced variations.

Most existing domain generalization methods struggle to bridge the synthetic-to-real gap under significant perspective changes. Mixing-based approaches exhibit severe performance degradation on both datasets. PolarMix suffers substantial drops on DALES (mIoU -22.95\%) and Toronto-3D (mIoU -7.42\%), accompanied by large-scale semantic confusion, particularly in the Vehicle category (Fig.~\ref{fig:T3D}(c), Fig.~\ref{fig:DALES}(c)). UniMix similarly degrades on DALES and exhibits fragmented predictions on Toronto-3D. DGLSS demonstrates comparatively more stable behavior, with minimal performance changes on Toronto-3D and slight improvements on DALES, suggesting that source-domain simulation combined with semantic consistency constraints can partially alleviate structural gaps, though its capacity remains limited. Despite leveraging unlabeled target data, the domain adaptation method CoSMix also fails to achieve promising results, as reported in Table~\ref{tab:stpls3d_results}. Overall, these findings highlight that explicitly modeling perspective-related structural variations is essential for robust cross-domain generalization, particularly when transferring from synthetic point clouds to real-world acquisitions under diverse viewpoints.

\subsection{Ablation Studies}

The proposed CVGC framework consists of two key components: Cross-view Geometric Augmentation (CGA) and Geometric Consistency Regularization (GCR), as introduced in Sections III-B and III-C, respectively. To assess the contribution of each component, we conduct ablation studies under both experimental groups described above.

\input{table/Ablation_H3D}

As shown in Tables~\ref{tab:h3d_ablation} and \ref{tab:stpls3d_ablation}, CGA plays a dominant role when the target domain exhibits severe density mismatch. For example, in the H3D-based setting, transferring to ISPRS Vaihingen yields a substantial improvement, with mIoU increasing from 31.20\% to 53.17\%. Similar trends are observed in the synthetic-to-real scenario using STPLS3D as the source, where CGA effectively mitigates large distribution gaps caused by cross-view density and structural differences. These results confirm that CGA primarily addresses domain shifts at the distribution level by simulating diverse density patterns and viewpoint-dependent missingness.

\input{table/Ablation_STPLS3D}

We further analyze the contribution of GCR by comparing the full model with the augmentation-only variant. When density discrepancies are relatively limited, such as transferring from H3D to Paris-Lille-3D, GCR provides significant additional gains (+5.83\% mIoU), demonstrating the importance of enforcing geometric consistency across views. In more challenging settings with strong density shifts, GCR still delivers consistent complementary improvements (e.g., +2.01\% mIoU on ISPRS Vaihingen), indicating its effectiveness in further regularizing geometric representations beyond data augmentation alone.

These results indicate that CGA primarily addresses domain gaps induced by density and visibility variations, while GCR operates at the representation level by aligning local structures across viewpoints. Acting at complementary levels, the two components jointly enhance cross-view generalization without requiring any target-domain data.

%% file: table/Method_H3D.tex
\begin{table*}[t]
\centering
\caption{Quantitative comparison on Hessigheim-3D (Source) to Paris-Lille-3D and ISPRS Vaihingen (Targets) cross-view benchmarks.}
\label{tab:h3d_results}
\setlength{\tabcolsep}{4pt} 
\renewcommand{\arraystretch}{1.0} 
\resizebox{0.9\textwidth}{!}{
\begin{tabular}{c c l ccccc c}
\toprule
\multirow{2}{*}{Source} &
\multirow{2}{*}{Target} &
\multirow{2}{*}{Method} &
\multicolumn{5}{c}{Per-class IoU (\%)} &
\multirow{2}{*}{mIoU} \\
\cmidrule(lr){4-8}
&  &  & Ground & Building & Natural & Vehicle & U.Furniture &  \\
\midrule
\multirow{10}{*}[-1.5em]{Hessigheim-3D}
 & \multirow{6}{*}{Paris-Lille-3D}
 & Source-only 
 & 91.98 & 83.83 & \underline{82.55} & 50.65 & 21.30 & 66.06 \\
 &  & CoSMix\cite{saltori2022cosmix} 
 & \textbf{95.29} & \underline{87.05} & 80.04 & \underline{64.71} & \textbf{31.93} & \underline{71.80} \\
 &  & PolarMix\cite{xiao2022polarmix} 
 & \underline{94.20} & 82.75 & 62.13 & 29.90 &  6.97 & 55.19 \\
 &  & DGLSS\cite{kim2023single} 
 & 92.33 & 84.89 & 81.12 & 42.58 & 20.23 & 64.23 \\
 &  & UniMix\cite{zhao2024unimix} 
 & 93.44 & 69.99 & 62.12 & 44.06 & \underline{25.04} & 58.93 \\
 &  & \textbf{Ours} 
 & 92.74 & \textbf{87.87} & \textbf{84.51} & \textbf{78.93} & 20.29 & \textbf{72.87} \\
\cmidrule(lr){2-9}
 &  \multirow{6}{*}{ISPRS Vaihingen}
 & Source-only 
 & 80.26 & 25.08 & 37.08 &  9.10 &  4.49 & 31.20 \\
 &  & CoSMix\cite{saltori2022cosmix} 
 & \underline{82.73} & 43.12 & 44.50 & 15.68 &  \underline{7.23} & \underline{38.65} \\
 &  & PolarMix\cite{xiao2022polarmix} 
 & 82.02 & 43.44 & \underline{47.50} & 10.26 &  3.48 & 37.34 \\
 &  & DGLSS\cite{kim2023single} 
 & 81.12 & 39.42 & 44.11 & 13.81 &  6.47 & 36.99 \\
 &  & UniMix\cite{zhao2024unimix} 
 & 73.54 & \underline{46.53} & 38.99 & \underline{15.90} &  3.32 & 35.65 \\
 &  & \textbf{Ours} 
 & \textbf{86.69} & \textbf{79.14} & \textbf{68.57} & \textbf{31.86} &  \textbf{9.62} & \textbf{55.18} \\
\bottomrule
\end{tabular}
}
\end{table*}

%% file: table/Method_STPLS3D.tex
\begin{table*}[t]
\centering
\caption{Quantitative comparison on STPLS3D (Source) to Toronto-3D and DALES (Targets) cross-view benchmarks.}
\label{tab:stpls3d_results}
\setlength{\tabcolsep}{4pt} 
\renewcommand{\arraystretch}{1.0} 
\resizebox{0.9\textwidth}{!}{
\begin{tabular}{c c l cccccc c}
\toprule
\multirow{2}{*}[-0.3em]{Source} &
\multirow{2}{*}[-0.3em]{Target} &
\multirow{2}{*}[-0.3em]{Method} &
\multicolumn{6}{c}{Per-class IoU (\%)} &
\multirow{2}{*}[-0.3em]{mIoU}\\
\cmidrule(lr){4-9}
 &  &  & Ground & Building & Natural & Vehicle & Pole & Fence & \\
\midrule
\multirow{10}{*}[-1.5em]{STPLS3D}
 & \multirow{6}{*}{Toronto-3D}
 & Source-only 
 & \textbf{96.51} & 57.83 & \underline{84.51} &  9.44 & \underline{35.72} &  2.70 & 47.79\\
 &  & CoSMix\cite{saltori2022cosmix} 
 & 89.73 & 17.96 & 32.43 &  6.80 & 15.34 &  0.93 & 27.20 \\
 &  & PolarMix\cite{xiao2022polarmix} 
 & 89.45 & 45.08 & 62.58 & 13.09 & 28.34 &  \underline{3.68} & 40.37 \\
 &  & DGLSS\cite{kim2023single} 
 & 96.13 & \underline{60.15} & 81.63 & 12.18 & \textbf{35.90} &  3.41 & 48.23 \\
 &  & UniMix\cite{zhao2024unimix} 
 & 95.85 & 41.87 & \textbf{85.19} & \underline{32.05} & 31.74 &  3.08 & \underline{48.30} \\
 &  & \textbf{Ours}  
 & \underline{96.35} & \textbf{67.35} & 81.24 & \textbf{34.72} & 20.62 &  \textbf{7.42} & \textbf{51.28} \\
\cmidrule(lr){2-10}
 &  \multirow{6}{*}{DALES}
 & Source-only 
 & 94.05 & 74.50 & 80.56 & \underline{34.14} & 33.53 & \underline{34.63} & 58.57 \\
 &  & CoSMix\cite{saltori2022cosmix} 
 & 93.36 & 81.23 & 77.18 & 22.30 & 10.26 & 12.85 & 49.53 \\
 &  & PolarMix\cite{xiao2022polarmix}  
 & 79.59 & 17.32 & 61.55 &  6.77 & 18.91 & 29.57 & 35.62 \\
 &  & DGLSS\cite{kim2023single}  
 & \textbf{94.29} & \underline{82.50} & \underline{83.47} & 22.79 & \underline{36.33} & 33.44 & \underline{58.80} \\
 &  & UniMix\cite{zhao2024unimix}  
 & 91.02 & 60.74 & 78.05 & 29.39 & 18.90 & 27.08 & 50.86 \\
 &  & \textbf{Ours} 
 & \underline{94.25} & \textbf{89.66} & \textbf{87.13} & \textbf{38.57} & \textbf{36.88} & \textbf{40.01} & \textbf{64.42} \\
\bottomrule
\end{tabular}
}
\end{table*}

%% file: table/Ablation_H3D.tex
\begin{table}[t]
\centering
\caption{Ablation study on Hessigheim-3D as source, with Paris-Lille-3D and ISPRS Vaihingen as targets.}
\label{tab:h3d_ablation}
\setlength{\tabcolsep}{2.5pt} 

\resizebox{0.9\columnwidth}{!}{
\begin{tabular}{c c c c c}
\toprule
\multirow{2}{*}[-0.3em]{Source} &
\multirow{2}{*}[-0.3em]{Target} & 
\multicolumn{2}{c}{Module} &
\multirow{2}{*}[-0.3em]{mIoU}\\
\cmidrule(lr){3-4}
 &  & CGA & GCR & \\
\midrule
\multirow{6}{*}[-0.25em]{Hessigheim-3D}
 & \multirow{3}{*}{Paris-Lille-3D}
 & \ding{55} &\ding{55}  & 66.06 \\
 &  & \ding{51} & \ding{55}  & 67.04 \\
 &  & \ding{51} & \ding{51} & \textbf{72.87} \\
\cmidrule(lr){2-5}
 & \multirow{3}{*}{ISPRS Vaihingen}
 & \ding{55} & \ding{55}    & 31.20 \\
 &  & \ding{51} &  \ding{55}          & 53.17 \\
 &  & \ding{51} & \ding{51} & \textbf{55.18} \\
\bottomrule
\end{tabular}
}
\end{table}

%% file: table/Ablation_STPLS3D.tex
\begin{table}[t]
\centering
\caption{Ablation study on STPLS3D as source, with Toronto-3D and DALES as targets.}
\label{tab:stpls3d_ablation}
\setlength{\tabcolsep}{5.5pt} 

\resizebox{0.9\columnwidth}{!}{
\begin{tabular}{c c c c c}
\toprule
\multirow{2}{*}[-0.3em]{Source} &
\multirow{2}{*}[-0.3em]{Target} & 
\multicolumn{2}{c}{Module} &
\multirow{2}{*}[-0.3em]{mIoU} \\
\cmidrule(lr){3-4}
 &  & CGA & GCR & \\
\midrule
\multirow{6}{*}[-0.25em]{STPLS3D}
 & \multirow{3}{*}{Toronto-3D}
 & \ding{55} &\ding{55}  & 47.79 \\
 &  & \ding{51} & \ding{55} & 50.42 \\
 &  & \ding{51} & \ding{51} & \textbf{51.28} \\
\cmidrule(lr){2-5}
 & \multirow{3}{*}{DALES}
 &\ding{55}  &\ding{55}  & 58.57 \\
 &  & \ding{51} &\ding{55}  & 61.91 \\
 &  & \ding{51} & \ding{51} & \textbf{64.42} \\
\bottomrule
\end{tabular}
}
\end{table}

%% file: sec/5_conclusions.tex
\section{Conclusions}

In this paper, we formulate for the first time the largely overlooked yet increasingly important problem of cross-view domain generalization for LiDAR point cloud semantic segmentation, and propose a novel framework CVGC to address this challenging problem. CVGC is built upon the key idea of generating multiple cross-view variants of the same scene through geometric augmentation, and enforcing invariance of semantic predictions and spatial occupancy across these geometrically transformed observations, enabling the model to learn view-invariant representations. Extensive experiments on six public LiDAR datasets spanning highly diverse acquisition viewpoints from airborne and UAV based to vehicle mounted platforms demonstrate that CVGC consistently outperforms existing domain generalization methods. Our study highlights the necessity of explicitly accounting for LiDAR acquisition viewpoints in domain generalization research and showcases the potential of domain-generalized models for robust large-scale LiDAR point cloud interpretation. Future work will extend geometric consistency learning beyond intra-scene augmentation toward cross-scene structural invariance, and incorporate richer sensor-aware modeling to further improve generalization under real-world cross-view conditions.

%% file: CrossPoints.bbl
\begin{thebibliography}{10}
\providecommand{\url}[1]{#1}
\csname url@samestyle\endcsname
\providecommand{\newblock}{\relax}
\providecommand{\bibinfo}[2]{#2}
\providecommand{\BIBentrySTDinterwordspacing}{\spaceskip=0pt\relax}
\providecommand{\BIBentryALTinterwordstretchfactor}{4}
\providecommand{\BIBentryALTinterwordspacing}{\spaceskip=\fontdimen2\font plus
\BIBentryALTinterwordstretchfactor\fontdimen3\font minus
  \fontdimen4\font\relax}
\providecommand{\BIBforeignlanguage}[2]{{%
\expandafter\ifx\csname l@#1\endcsname\relax
\typeout{** WARNING: IEEEtran.bst: No hyphenation pattern has been}%
\typeout{** loaded for the language `#1'. Using the pattern for}%
\typeout{** the default language instead.}%
\else
\language=\csname l@#1\endcsname
\fi
#2}}
\providecommand{\BIBdecl}{\relax}
\BIBdecl

\bibitem{geiger2012we}
A.~Geiger, P.~Lenz, and R.~Urtasun, ``Are we ready for autonomous driving? the
  kitti vision benchmark suite,'' in \emph{2012 IEEE Conference on Computer
  Vision and Pattern Recognition}, 2012, pp. 3354--3361.

\bibitem{lafarge2012creating}
F.~Lafarge and C.~Mallet, ``Creating large-scale city models from 3d-point
  clouds: a robust approach with hybrid representation,'' \emph{International
  journal of computer vision}, vol.~99, no.~1, pp. 69--85, 2012.

\bibitem{pan2024pin}
Y.~Pan, X.~Zhong, L.~Wiesmann, T.~Posewsky, J.~Behley, and C.~Stachniss,
  ``Pin-slam: Lidar slam using a point-based implicit neural representation for
  achieving global map consistency,'' \emph{IEEE Transactions on Robotics},
  vol.~40, pp. 4045--4064, 2024.

\bibitem{gao2025lidar}
Y.~Gao, S.~Xia, P.~Wang, X.~Xi, S.~Nie, and C.~Wang, ``Lidar remote sensing
  meets weak supervision: Concepts, methods, and perspectives,'' \emph{arXiv
  preprint arXiv:2503.18384}, 2025.

\bibitem{choy20194d}
C.~Choy, J.~Gwak, and S.~Savarese, ``4d spatio-temporal convnets: Minkowski
  convolutional neural networks,'' in \emph{Proceedings of the IEEE/CVF
  Conference on Computer Vision and Pattern Recognition (CVPR)}, June 2019.

\bibitem{guo2020deep}
Y.~Guo, H.~Wang, Q.~Hu, H.~Liu, L.~Liu, and M.~Bennamoun, ``Deep learning for
  3d point clouds: A survey,'' \emph{IEEE Transactions on Pattern Analysis and
  Machine Intelligence}, vol.~43, no.~12, pp. 4338--4364, 2021.

\bibitem{xu2025frnet}
X.~Xu, L.~Kong, H.~Shuai, and Q.~Liu, ``Frnet: Frustum-range networks for
  scalable lidar segmentation,'' \emph{IEEE Transactions on Image Processing},
  vol.~34, pp. 2173--2186, 2025.

\bibitem{zhou2022domain}
K.~Zhou, Z.~Liu, Y.~Qiao, T.~Xiang, and C.~C. Loy, ``Domain generalization: A
  survey,'' \emph{IEEE Transactions on Pattern Analysis and Machine
  Intelligence}, vol.~45, no.~4, pp. 4396--4415, 2023.

\bibitem{wang2022generalizing}
J.~Wang, C.~Lan, C.~Liu, Y.~Ouyang, T.~Qin, W.~Lu, Y.~Chen, W.~Zeng, and P.~S.
  Yu, ``Generalizing to unseen domains: A survey on domain generalization,''
  \emph{IEEE Transactions on Knowledge and Data Engineering}, vol.~35, no.~8,
  pp. 8052--8072, 2023.

\bibitem{kim2023single}
H.~Kim, Y.~Kang, C.~Oh, and K.-J. Yoon, ``Single domain generalization for
  lidar semantic segmentation,'' in \emph{Proceedings of the IEEE/CVF
  Conference on Computer Vision and Pattern Recognition (CVPR)}, June 2023, pp.
  17\,587--17\,598.

\bibitem{zhao2024unimix}
H.~Zhao, J.~Zhang, Z.~Chen, S.~Zhao, and D.~Tao, ``Unimix: Towards domain
  adaptive and generalizable lidar semantic segmentation in adverse weather,''
  in \emph{Proceedings of the IEEE/CVF Conference on Computer Vision and
  Pattern Recognition (CVPR)}, June 2024, pp. 14\,781--14\,791.

\bibitem{shi2025l2rsi}
Z.~Shi, X.~Zhang, W.~Xu, Y.~Xia, Y.~Zang, S.~Shen, and C.~Wang, ``L2rsi:
  Cross-view lidar-based place recognition for large-scale urban scenes via
  remote sensing imagery,'' \emph{arXiv preprint arXiv:2503.11245}, 2025.

\bibitem{10657359}
J.~Ye, Q.~Luo, J.~Yu, H.~Zhong, Z.~Zheng, C.~He, and W.~Li, ``Sg-bev:
  Satellite-guided bev fusion for cross-view semantic segmentation,'' in
  \emph{2024 IEEE/CVF Conference on Computer Vision and Pattern Recognition
  (CVPR)}, 2024, pp. 27\,748--27\,757.

\bibitem{10373898}
Z.~Xia, O.~Booij, and J.~F.~P. Kooij, ``Convolutional cross-view pose
  estimation,'' \emph{IEEE Transactions on Pattern Analysis and Machine
  Intelligence}, vol.~46, no.~5, pp. 3813--3831, 2024.

\bibitem{Liang_2025_ICCV}
A.~Liang, L.~Kong, D.~Lu, Y.~Liu, J.~Fang, H.~Zhao, and W.~T. Ooi,
  ``Perspective-invariant 3d object detection,'' in \emph{Proceedings of the
  IEEE/CVF International Conference on Computer Vision (ICCV)}, October 2025,
  pp. 27\,725--27\,738.

\bibitem{ren2022adela}
H.~Ren, Y.~Yang, H.~Wang, B.~Shen, Q.~Fan, Y.~Zheng, C.~K. Liu, and L.~J.
  Guibas, ``Adela: Automatic dense labeling with attention for viewpoint shift
  in semantic segmentation,'' in \emph{Proceedings of the IEEE/CVF Conference
  on Computer Vision and Pattern Recognition (CVPR)}, June 2022, pp.
  8079--8089.

\bibitem{truong2024eagle}
T.-D. Truong, U.~Prabhu, D.~Wang, B.~Raj, S.~Gauch, J.~Subbiah, and K.~Luu,
  ``Eagle: Efficient adaptive geometry-based learning in cross-view
  understanding,'' in \emph{Advances in Neural Information Processing Systems},
  vol.~37.\hskip 1em plus 0.5em minus 0.4em\relax Curran Associates, Inc.,
  2024, pp. 137\,309--137\,333.

\bibitem{coors2019nova}
B.~Coors, A.~P. Condurache, and A.~Geiger, ``Nova: Learning to see in novel
  viewpoints and domains,'' in \emph{2019 International Conference on 3D Vision
  (3DV)}, 2019, pp. 116--125.

\bibitem{klinghoffer2023towards}
T.~Klinghoffer, J.~Philion, W.~Chen, O.~Litany, Z.~Gojcic, J.~Joo, R.~Raskar,
  S.~Fidler, and J.~M. Alvarez, ``Towards viewpoint robustness in bird's eye
  view segmentation,'' in \emph{Proceedings of the IEEE/CVF International
  Conference on Computer Vision (ICCV)}, October 2023, pp. 8515--8524.

\bibitem{varney2020dales}
N.~Varney, V.~K. Asari, and Q.~Graehling, ``Dales: A large-scale aerial lidar
  data set for semantic segmentation,'' in \emph{Proceedings of the IEEE/CVF
  Conference on Computer Vision and Pattern Recognition (CVPR) Workshops}, June
  2020.

\bibitem{rist2019cross}
C.~B. Rist, M.~Enzweiler, and D.~M. Gavrila, ``Cross-sensor deep domain
  adaptation for lidar detection and segmentation,'' in \emph{2019 IEEE
  Intelligent Vehicles Symposium (IV)}, 2019, pp. 1535--1542.

\bibitem{xiao2022transfer}
A.~Xiao, J.~Huang, D.~Guan, F.~Zhan, and S.~Lu, ``Transfer learning from
  synthetic to real lidar point cloud for semantic segmentation,''
  \emph{Proceedings of the AAAI Conference on Artificial Intelligence},
  vol.~36, no.~3, pp. 2795--2803, Jun. 2022.

\bibitem{cosmix2023}
C.~Saltori, F.~Galasso, G.~Fiameni, N.~Sebe, F.~Poiesi, and E.~Ricci,
  ``Compositional semantic mix for domain adaptation in point cloud
  segmentation,'' \emph{IEEE Transactions on Pattern Analysis and Machine
  Intelligence}, vol.~45, no.~12, pp. 14\,234--14\,247, 2023.

\bibitem{shaban2023lidar}
A.~Shaban, J.~Lee, S.~Jung, X.~Meng, and B.~Boots, ``Lidar-uda: Self-ensembling
  through time for unsupervised lidar domain adaptation,'' in \emph{Proceedings
  of the IEEE/CVF International Conference on Computer Vision (ICCV)}, October
  2023, pp. 19\,784--19\,794.

\bibitem{PrototypeUDA2023}
Z.~Yuan, M.~Cheng, W.~Zeng, Y.~Su, W.~Liu, S.~Yu, and C.~Wang,
  ``Prototype-guided multitask adversarial network for cross-domain lidar point
  clouds semantic segmentation,'' \emph{IEEE Transactions on Geoscience and
  Remote Sensing}, vol.~61, pp. 1--13, 2023.

\bibitem{chen2024bridging}
S.~Chen, B.~Yang, Y.~Xia, M.~Cheng, S.~Shen, and C.~Wang, ``Bridging lidar
  gaps: a multi-lidars domain adaptation dataset for 3d semantic
  segmentation,'' in \emph{Proceedings of the Thirty-Third International Joint
  Conference on Artificial Intelligence}, ser. IJCAI '24, 2024.

\bibitem{luo2025cross}
H.~Luo, Z.~Chen, F.~Ye, T.~Huang, H.~He, and W.~Hu, ``Cross-sensor adaptive
  semantic segmentation for mobile laser scanning point clouds based on
  continuous potential scene surface reconstruction,'' \emph{ISPRS Journal of
  Photogrammetry and Remote Sensing}, vol. 228, pp. 537--551, 2025.

\bibitem{xiao2022polarmix}
A.~Xiao, J.~Huang, D.~Guan, K.~Cui, S.~Lu, and L.~Shao, ``Polarmix: A general
  data augmentation technique for lidar point clouds,'' in \emph{Advances in
  Neural Information Processing Systems}, S.~Koyejo, S.~Mohamed, A.~Agarwal,
  D.~Belgrave, K.~Cho, and A.~Oh, Eds., vol.~35.\hskip 1em plus 0.5em minus
  0.4em\relax Curran Associates, Inc., 2022, pp. 11\,035--11\,048.

\bibitem{xiao2023domain}
A.~Xiao, D.~Guan, X.~Zhang, and S.~Lu, ``Domain adaptive lidar point cloud
  segmentation with 3d spatial consistency,'' \emph{IEEE Transactions on
  Multimedia}, vol.~26, pp. 5536--5547, 2024.

\bibitem{saltori2022cosmix}
C.~Saltori, F.~Galasso, G.~Fiameni, N.~Sebe, E.~Ricci, and F.~Poiesi, ``Cosmix:
  Compositional semantic mix for domain adaptation in 3d lidar
  segmentation,'' in \emph{Computer Vision -- ECCV 2022}.\hskip 1em plus 0.5em
  minus 0.4em\relax Cham: Springer Nature Switzerland, 2022, pp. 586--602.

\bibitem{kong2023lasermix}
L.~Kong, J.~Ren, L.~Pan, and Z.~Liu, ``Lasermix for semi-supervised lidar
  semantic segmentation,'' in \emph{Proceedings of the IEEE/CVF Conference on
  Computer Vision and Pattern Recognition (CVPR)}, June 2023, pp.
  21\,705--21\,715.

\bibitem{Yuan_2024_CVPR}
Z.~Yuan, W.~Zeng, Y.~Su, W.~Liu, M.~Cheng, Y.~Guo, and C.~Wang,
  ``Density-guided translator boosts synthetic-to-real unsupervised domain
  adaptive segmentation of 3d point clouds,'' in \emph{Proceedings of the
  IEEE/CVF Conference on Computer Vision and Pattern Recognition (CVPR)}, June
  2024, pp. 23\,303--23\,312.

\bibitem{xiao2024domain}
A.~Xiao, J.~Huang, K.~Liu, D.~Guan, X.~Zhang, and S.~Lu, ``Domain adaptive
  lidar point cloud segmentation via density-aware self-training,'' \emph{IEEE
  Transactions on Intelligent Transportation Systems}, vol.~25, no.~10, pp.
  13\,627--13\,639, 2024.

\bibitem{luo2020unsupervised}
H.~Luo, K.~Khoshelham, L.~Fang, and C.~Chen, ``Unsupervised scene adaptation
  for semantic segmentation of urban mobile laser scanning point clouds,''
  \emph{ISPRS Journal of Photogrammetry and Remote Sensing}, vol. 169, pp.
  253--267, 2020.

\bibitem{yi2021complete}
L.~Yi, B.~Gong, and T.~Funkhouser, ``Complete \& label: A domain adaptation
  approach to semantic segmentation of lidar point clouds,'' in
  \emph{Proceedings of the IEEE/CVF Conference on Computer Vision and Pattern
  Recognition (CVPR)}, June 2021, pp. 15\,363--15\,373.

\bibitem{boulch2023also}
A.~Boulch, C.~Sautier, B.~Michele, G.~Puy, and R.~Marlet, ``Also: Automotive
  lidar self-supervision by occupancy estimation,'' in \emph{Proceedings of the
  IEEE/CVF Conference on Computer Vision and Pattern Recognition (CVPR)}, June
  2023, pp. 13\,455--13\,465.

\bibitem{SALUDA2024}
B.~Michele, A.~Boulch, G.~Puy, T.-H. Vu, R.~Marlet, and N.~Courty, ``Saluda:
  Surface-based automotive lidar unsupervised domain adaptation,'' in
  \emph{2024 International Conference on 3D Vision (3DV)}, 2024, pp. 421--431.

\bibitem{lao2024lit}
Y.~Lao, T.~Tang, X.~Wu, P.~Chen, K.~Yu, and H.~Zhao, ``Lit: Unifying lidar
  "languages" with lidar translator,'' in \emph{Advances in Neural Information
  Processing Systems}, A.~Globerson, L.~Mackey, D.~Belgrave, A.~Fan, U.~Paquet,
  J.~Tomczak, and C.~Zhang, Eds., vol.~37.\hskip 1em plus 0.5em minus
  0.4em\relax Curran Associates, Inc., 2024, pp. 93\,767--93\,789.

\bibitem{li2026its}
B.~Li, Y.~Pang, D.~Kükenbrink, L.~Wang, D.~Kong, and M.~Marty, ``Its-net: A
  platform and sensor agnostic 3d deep learning model for individual tree
  segmentation using aerial lidar data,'' \emph{ISPRS Journal of Photogrammetry
  and Remote Sensing}, vol. 231, pp. 719--744, 2026.

\bibitem{WANG2025422}
P.~Wang, W.~Yao, J.~Shao, and Z.~He, ``Test-time adaptation for geospatial
  point cloud semantic segmentation with distinct domain shifts,'' \emph{ISPRS
  Journal of Photogrammetry and Remote Sensing}, vol. 229, pp. 422--435, 2025.

\bibitem{whu3d}
X.~Han, C.~Liu, Y.~Zhou, K.~Tan, Z.~Dong, and B.~Yang, ``Whu-urban3d: An urban
  scene lidar point cloud dataset for semantic instance segmentation,''
  \emph{ISPRS Journal of Photogrammetry and Remote Sensing}, vol. 209, pp.
  500--513, 2024.

\bibitem{zhang2018mixup}
H.~Zhang, M.~Cisse, Y.~N. Dauphin, and D.~Lopez-Paz, ``mixup: Beyond empirical
  risk minimization,'' in \emph{Proceedings of the International Conference on
  Learning Representations (ICLR)}.\hskip 1em plus 0.5em minus 0.4em\relax
  Vancouver, Canada: OpenReview.net, 2018.

\bibitem{li2018domain}
H.~Li, S.~J. Pan, S.~Wang, and A.~C. Kot, ``Domain generalization with
  adversarial feature learning,'' in \emph{Proceedings of the IEEE Conference
  on Computer Vision and Pattern Recognition (CVPR)}, June 2018, pp.
  5400--5409.

\bibitem{li2018learning}
D.~Li, Y.~Yang, Y.-Z. Song, and T.~Hospedales, ``Learning to generalize:
  Meta-learning for domain generalization,'' \emph{Proceedings of the AAAI
  Conference on Artificial Intelligence}, vol.~32, no.~1, April 2018.

\bibitem{espadinha2021lidar}
J.~Espadinha, I.~Lebedev, L.~Lukic, and A.~Bernardino, ``Lidar data noise
  models and methodology for sim-to-real domain generalization and adaptation
  in autonomous driving perception,'' in \emph{2021 IEEE Intelligent Vehicles
  Symposium (IV)}, 2021, pp. 797--803.

\bibitem{xiao20233d}
A.~Xiao, J.~Huang, W.~Xuan, R.~Ren, K.~Liu, D.~Guan, A.~El~Saddik, S.~Lu, and
  E.~P. Xing, ``3d semantic segmentation in the wild: Learning generalized
  models for adverse-condition point clouds,'' in \emph{Proceedings of the
  IEEE/CVF Conference on Computer Vision and Pattern Recognition (CVPR)}, June
  2023, pp. 9382--9392.

\bibitem{Sanchez_2023_ICCV}
J.~Sanchez, J.-E. Deschaud, and F.~Goulette, ``Domain generalization of 3d
  semantic segmentation in autonomous driving,'' in \emph{Proceedings of the
  IEEE/CVF International Conference on Computer Vision (ICCV)}, October 2023,
  pp. 18\,077--18\,087.

\bibitem{kim2024rethinking}
J.~Kim, J.~Woo, J.~Kim, and S.~Im, ``Rethinking lidar domain generalization:
  Single source as multiple density domains,'' in \emph{Computer Vision --
  ECCV 2024}.\hskip 1em plus 0.5em minus 0.4em\relax Cham: Springer Nature
  Switzerland, 2025, pp. 310--327.

\bibitem{kim2024density}
J.~Kim, J.~Woo, U.~Shin, J.~Oh, and S.~Im, ``Density-aware domain
  generalization for lidar semantic segmentation,'' in \emph{2024 IEEE/RSJ
  International Conference on Intelligent Robots and Systems (IROS)}, 2024, pp.
  9573--9580.

\bibitem{COLA_2025_Sanchez}
J.~Sanchez, J.-E. Deschaud, and F.~Goulette, ``Cola: Coarse-label multisource
  lidar semantic segmentation for autonomous driving,'' \emph{IEEE Transactions
  on Robotics}, vol.~41, pp. 1742--1754, 2025.

\bibitem{he2024domain}
P.~He, L.~Jiao, L.~Li, X.~Liu, F.~Liu, W.~Ma, S.~Yang, and R.~Shang, ``Domain
  generalization-aware uncertainty introspective learning for 3d point clouds
  segmentation,'' in \emph{Proceedings of the 32nd ACM International Conference
  on Multimedia}, ser. MM '24.\hskip 1em plus 0.5em minus 0.4em\relax New York,
  NY, USA: Association for Computing Machinery, 2024, p. 651–660.

\bibitem{he2025domain}
P.~He, L.~Li, L.~Jiao, R.~Shang, F.~Liu, S.~Wang, X.~Liu, and W.~Ma,
  ``Domain-aware category-level geometry learning segmentation for 3d point
  clouds,'' in \emph{Proceedings of the IEEE/CVF International Conference on
  Computer Vision (ICCV)}, October 2025, pp. 28\,324--28\,333.

\bibitem{li2023bev}
M.~Li, Y.~Zhang, X.~Ma, Y.~Qu, and Y.~Fu, ``Bev-dg: Cross-modal learning under
  bird's-eye view for domain generalization of 3d semantic segmentation,'' in
  \emph{Proceedings of the IEEE/CVF International Conference on Computer Vision
  (ICCV)}, October 2023, pp. 11\,632--11\,642.

\bibitem{hegde2025multimodal}
D.~Hegde, S.~Lohit, K.-C. Peng, M.~Jones, and V.~Patel, ``Multimodal 3d object
  detection on unseen domains,'' in \emph{Proceedings of the IEEE/CVF
  Conference on Computer Vision and Pattern Recognition (CVPR) Workshops}, June
  2025, pp. 2524--2534.

\bibitem{saltori2023walking}
C.~Saltori, A.~Osep, E.~Ricci, and L.~Leal-Taix\'e, ``Walking your lidog: A
  journey through multiple domains for lidar semantic segmentation,'' in
  \emph{Proceedings of the IEEE/CVF International Conference on Computer Vision
  (ICCV)}, October 2023, pp. 196--206.

\bibitem{kolle2021hessigheim}
M.~Kölle, D.~Laupheimer, S.~Schmohl, N.~Haala, F.~Rottensteiner, J.~D. Wegner,
  and H.~Ledoux, ``The hessigheim 3d (h3d) benchmark on semantic segmentation
  of high-resolution 3d point clouds and textured meshes from uav lidar and
  multi-view-stereo,'' \emph{ISPRS Open Journal of Photogrammetry and Remote
  Sensing}, vol.~1, p. 100001, 2021.

\bibitem{roynard2018paris}
X.~Roynard, J.-E. Deschaud, and F.~Goulette, ``Paris-lille-3d: A large and
  high-quality ground-truth urban point cloud dataset for automatic
  segmentation and classification,'' \emph{The International Journal of
  Robotics Research}, vol.~37, no.~6, pp. 545--557, 2018.

\bibitem{rottensteiner2012isprs}
F.~Rottensteiner, G.~Sohn, J.~Jung, M.~Gerke, C.~Baillard, S.~Benitez, and
  U.~Breitkopf, ``The isprs benchmark on urban object classification and 3d
  building reconstruction,'' in \emph{ISPRS Annals of the Photogrammetry,
  Remote Sensing and Spatial Information Sciences}, ser. Volume I-3, 2012, pp.
  293--298, iSPRS Test Project on Urban Classification and 3D Reconstruction
  Benchmark.

\bibitem{chen2022stpls3d}
M.~Chen, Q.~Hu, Z.~Yu, H.~THOMAS, A.~Feng, Y.~Hou, K.~McCullough, F.~Ren, and
  L.~Soibelman, ``Stpls3d: A large-scale synthetic and real aerial
  photogrammetry 3d point cloud dataset,'' in \emph{33rd British Machine Vision
  Conference 2022, {BMVC} 2022, London, UK, November 21-24, 2022}.\hskip 1em
  plus 0.5em minus 0.4em\relax {BMVA} Press, 2022.

\bibitem{Tan2020toronto}
W.~Tan, N.~Qin, L.~Ma, Y.~Li, J.~Du, G.~Cai, K.~Yang, and J.~Li, ``Toronto-3d:
  A large-scale mobile lidar dataset for semantic segmentation of urban
  roadways,'' in \emph{Proceedings of the IEEE/CVF Conference on Computer
  Vision and Pattern Recognition (CVPR) Workshops}, June 2020.

\end{thebibliography}
